\documentclass[10pt,twocolumn,letterpaper]{article}

\usepackage{cvpr}              
\definecolor{cvprblue}{rgb}{0.21,0.49,0.74}
\usepackage[pagebackref,breaklinks,colorlinks,allcolors=cvprblue]{hyperref}

\def\eqref#1{Eq.~(\ref{#1})}

\usepackage[ruled, nosemicolon]{algorithm2e}
\usepackage{changepage}
\usepackage{hyperref}
\usepackage{url}            
\usepackage{booktabs}       
\usepackage{amsfonts}       
\usepackage{nicefrac}       
\usepackage{microtype}      
\usepackage{colortbl}
\usepackage{graphicx}
\usepackage{amsmath}
\usepackage{amssymb}
\usepackage{mathtools}
\usepackage{makecell}
\usepackage{wrapfig}
\usepackage{media9}
\usepackage{caption}
\usepackage{graphicx}
\usepackage{placeins}
\usepackage{multimedia}
\usepackage{media9}
\usepackage{animate}
\usepackage{rotating}  
\usepackage{tikz}
\usepackage{hyperref}
\usepackage{needspace}  
\usepackage{pifont}
\usepackage{float}
\usepackage{svg}
\usepackage{adjustbox}

\usepackage{setspace} 
\usetikzlibrary{tikzmark}
\SetKwBlock{Stage}{\textbf{Stage}}{}
\usepackage{makecell}  
\definecolor{darkgreen}{RGB}{0,100,0}
\newcommand{\best}[1]{\cellcolor{blue!30}\textbf{#1}}
\newcommand{\second}[1]{\cellcolor{blue!15}\underline{#1}}

\def\paperID{20025} 
\def\confName{CVPR}
\def\confYear{2026}

\title{Zero4D: Training-Free 4D Video Generation From Single Video Using Off-the-Shelf Video Diffusion Models}

\author{
Jangho Park \qquad
Taesung Kwon \qquad
Jong Chul Ye\\
KAIST\\
{\tt\small \{jhq1234, star.kwon, jong.ye\}@kaist.ac.kr}
}

\begin{document}

\twocolumn[{%
\renewcommand\twocolumn[1][]{#1}%
\maketitle
\begin{center}
    \centering
    \captionsetup{type=figure}
    \includegraphics[width=\linewidth]{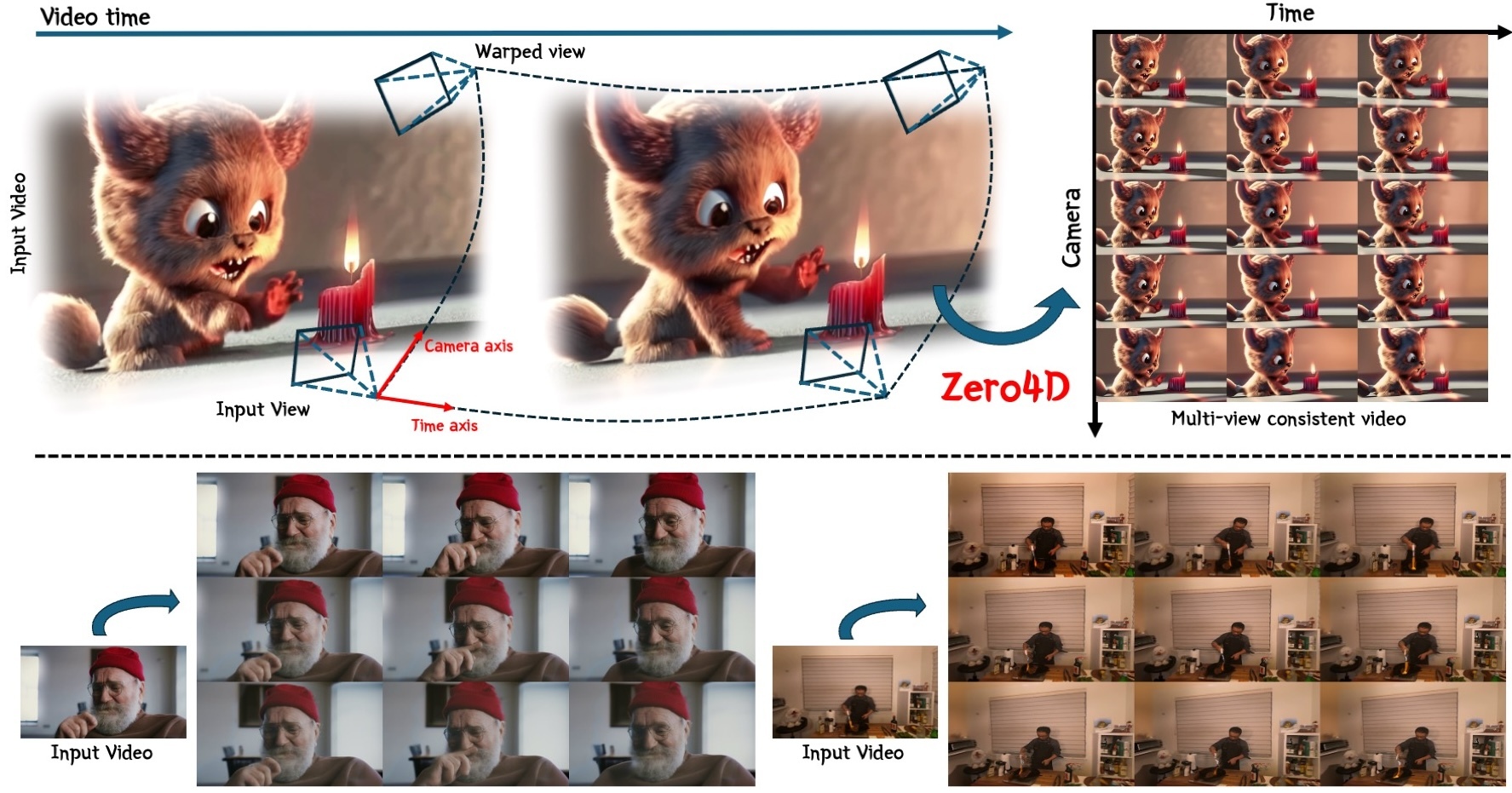}
\caption{\textbf{Zero4D} is a \textbf{training-free} multi-view synchronized video generation framework that takes a single monocular video and generates a grid of camera-time consistent frames. It first utilizes a depth estimation model to warp target view frames from the input video (top-left), then repurposes the image-to-video diffusion model to sample multi-view frames synchronized in both camera and temporal dimensions (top-right). Using an off-the-shelf video diffusion model without training, our approach can generate multi-view videos for both synthesized and real-world footage. Video results are available on our project page: \href{https://zero4dvid.github.io/}{\textit{zero4dvid.github.io.}}}
\label{fig:main}
\end{center}}]

\begin{abstract}
Multi-view and 4D video generation have emerged as important topics in generative modeling. However, existing approaches face key limitations: they often require orchestrating multiple video diffusion models with additional training, or involve computationally intensive training of full 4D diffusion models—despite limited availability of real-world 4D datasets. In this work, we propose a novel training-free 4D video generation method that leverages off-the-shelf video diffusion models to synthesize multi-view videos from a single input video. Our approach consists of two stages. First, we designate the edge frames in a spatio-temporal sampling grid as key frames and synthesize them using a video diffusion model, guided by depth-based warping to preserve structural and temporal consistency. Second, we interpolate the remaining frames to complete the spatio-temporal grid, again using a video diffusion model to maintain coherence. This two-step framework allows us to extend a single-view video into a multi-view 4D representation along novel camera trajectories, while maintaining spatio-temporal fidelity. Our method is training-free, requires no access to multi-view data, and fully utilizes existing generative video models offering a practical and effective solution for 4D video generation. 
\end{abstract}

\vspace{-1cm}

\section{Introduction}
Since the introduction of the diffusion and foundation models ~\cite{ddpm, sd, sv4d}, 3D reconstruction has advanced significantly, leading to unprecedented progress in representing the real world in 3D models. Combined with generative models, this success drives a renaissance in 3D generation, enabling more diverse and realistic content creation. These advancements extend beyond static scene or object reconstruction and generation, evolving toward dynamic 3D reconstruction and generation that aims to capture the real world. Previous works \citep{4dfy, stag4d, text-to-4d, animate124, tc4d} leverage video diffusion models (VDM) and Score Distillation Sampling (SDS) to enable dynamic 3D generation. However, most existing approaches primarily focus on generating dynamic objects in blank or simplified backgrounds (e.g., text-to-4D generation), leaving the more challenging task of reconstructing or generating real-world scenes from text prompts, reference images, or input videos largely unaddressed. In contrast to the abundance of high-quality datasets for 3D and video tasks, 4D datasets with multiview, temporally synchronized video remain extremely scarce. As a result, a core challenge in training 4D generative models for real-world scenes lies in the lack of comprehensive, large-scale multi-view video datasets. To overcome these limitations, recent works such as 4DiM \citep{4dim} propose a joint training diffusion model with 3D and video with a scarce 4D dataset. 
CAT4D \citep{cat4d} proposes training multi-view video diffusion models by curating a diverse collection of synthetic 4D data, 3D datasets, and monocular video sources.
DimensionX \citep{dimensionx} trains the spatial-temporal diffusion model independently with multiple LoRA, achieving multi-view videos via an additional refinement process.
Despite several approaches, the scarcity of high-quality 4D data makes it difficult to generalize to complex real-world scenes and poses fundamental challenges in training large multi-view video models.

To address these challenges, we introduce {\em Zero4D}—a novel zero-shot framework for 4D video generation. 
Zero4D generates synchronized multi-view 4D video from a single monocular input video by leveraging an off-the-shelf video diffusion model \citep{svd}, without requiring any additional training.
Building upon the prior observations \citep{4real-video, cat4d} that 4D video is composed of multiple video frames arranged along the spatio-temporal sampling grid (i.e., camera view and time axes),  generating a 4D video can be regarded as populating the sampling grid with consistent spatio-temporal frames. Consequently, our approach achieves this through two key steps:
(1) We first designate the boundary frames of the spatio-temporal sampling grid as key frames and synthesize them using a video diffusion model. To ensure structural fidelity, we incorporate a depth-based warping technique as guidance, encouraging the generated frames to conform to the underlying scene geometry. (2) We repurpose the interpolation capabilities of a video diffusion model to fill in the remaining frames through bidirectional diffusion sampling, resulting in a fully populated and temporally coherent 4D grid. Throughout both stages, our method enforces spatial and temporal consistency across the entire grid.

Our main contributions can be summarized as follows:
\begin{itemize}
    \item We propose a novel framework that can generate 4D video from a single video via an off-the-shelf video diffusion model without any training or large-scale datasets. To the best of our knowledge, our approach is the first interpolation based {\em training-free} method to generate synchronized multi-view video—  previously regarded as infeasible.
    \item This is made possible by a novel synchronization mechanism, which guarantees high-quality outputs while maintaining global spatio-temporal consistency. Specifically, we alternate bidirectional video interpolation across both the camera and temporal axes to align motion and appearance throughout the sequence.
    
    \item Our framework outperforms baselines in maintaining global spatio-temporal consistency and demonstrates robust 4D video generation capability, achieving competitive performance across diverse quantitative and qualitative evaluations even without additional training.
    
\end{itemize}


\begin{figure*}[!t]
    \centering
    \includegraphics[width=1.0\linewidth]{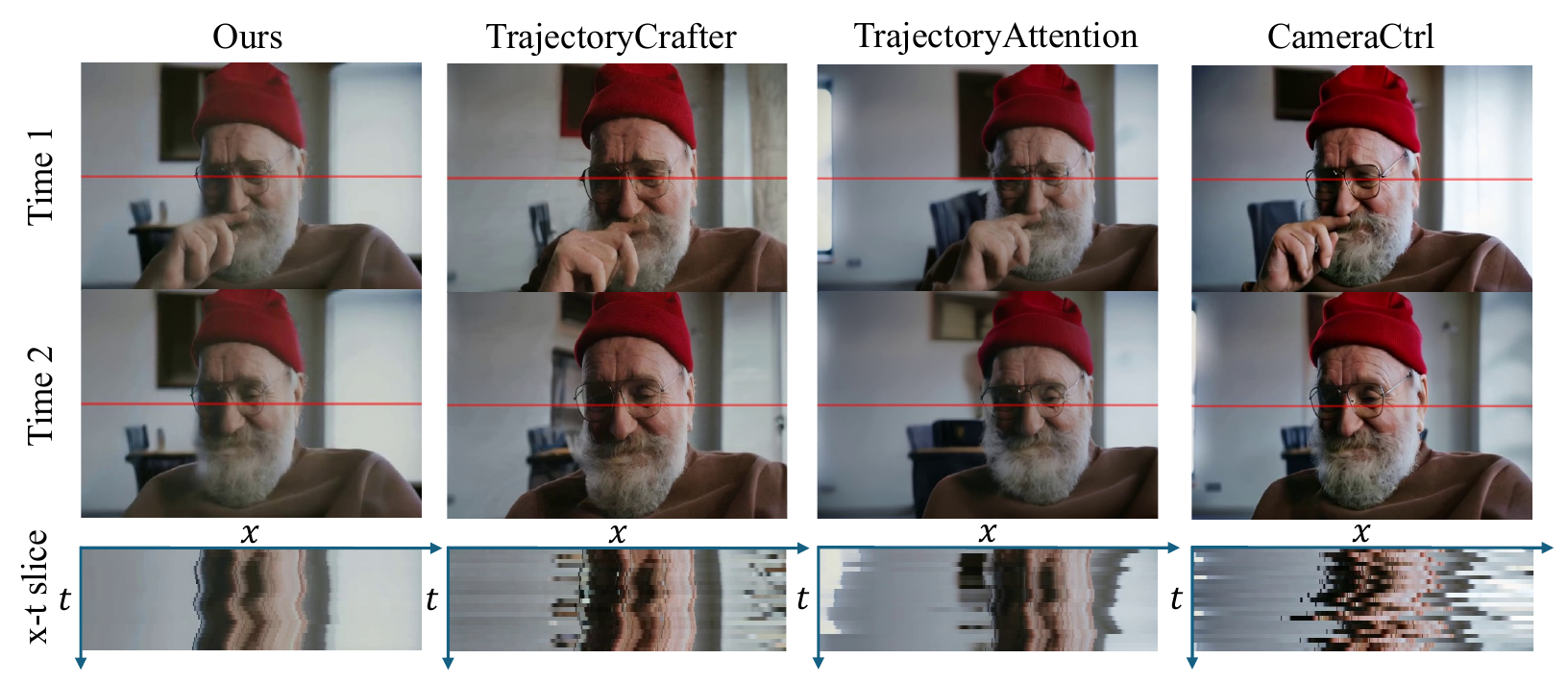}
    \caption{\textbf{Qualitative comparison.} 
    We compare our method with baseline models in terms of novel-view video generation and global spatio-temporal consistency. Given a single input video, both baselines and ours generate outputs across multiple views and time steps. To evaluate global consistency, we leverage baselines to produce bullet-time videos at all input frames and re-align them to a fixed viewpoint. We also visualize x–t slices (red lines) to highlight temporal coherence. While baselines exhibit inconsistencies across views and time, our method preserves spatio-temporal coherence and yields high-quality multi-view videos.}
    \label{fig:qualitative comparison}
\end{figure*}

\begin{table}[t]
\centering
\caption{Comparison of camera controllable VDM and 4D VDM. Unlike prior approaches, Zero4D can generate 4D-consistent videos with camera control without requiring additional training.}
\label{tab:model_comparison}

\begin{adjustbox}{max width=\linewidth}
\begin{tabular}{l|ccc}
    \toprule
    Model & \makecell{Training-Free} & \makecell{Camera Control} & \makecell{4D Consistency} \\
    \midrule
    Camera VDM \cite{trajcrafter,trajattn,cameractrl} 
    & \textcolor{red}{\ding{55}} 
    & \textcolor{darkgreen}{\ding{51}} 
    & \textcolor{red}{\ding{55}} \\
    
    4D VDM \cite{cat4d,dimensionx, sv4d, 4real-video}
    & \textcolor{red}{\ding{55}} 
    & \textcolor{darkgreen}{\ding{51}} 
    & \textcolor{darkgreen}{\ding{51}} \\
    
    \textbf{Zero4D (Ours)}
    & \textcolor{darkgreen}{\ding{51}} 
    & \textcolor{darkgreen}{\ding{51}} 
    & \textcolor{darkgreen}{\ding{51}} \\
    \bottomrule
\end{tabular}
\end{adjustbox}

\end{table}

\section{Related work}
\label{gen_inst}
\noindent\textbf{Video generation with camera control.}
Several studies try to train a multi-view diffusion model for spatially consistent image generation \citep{mvdream, imagedream, zero123, SPAD, cat3d, im3d}. ReCapture \citep{recapture} trains the novel camera trajectory video diffusion model from a single reference video with existing scene motion. They train LoRA layers with camera labels to regenerate the anchor video into a novel view. Camco\cite{camco} fintunes pre-trained video diffusion model with injecting Plücker embedding vector into a specific layer in the model. CameraCtrl \citep{cameractrl} proposes a plug-and-play camera module in the video diffusion model to control video generation with precise and smooth camera viewpoints. TrajectoryCrafter \citep{trajcrafter} and TrajectoryAttention \citep{trajattn} fine-tune video diffusion models to generate novel-view videos along a given camera trajectory using depth-based warping. These approaches can be categorized as \textit{camera-controllable video diffusion models}. However, although these models can synthesize novel views conditioned on warped videos, they fail to produce 4D-consistent videos that ensure global consistency across multiple views and multiple time steps (see Table~\ref{tab:model_comparison}).

\noindent\textbf{4D generation.}
Recent advancements in 4D generation have been driven by numerous pioneering works exploring various conditioning methods. Several approaches have leveraged score distillation sampling in conjunction with video diffusion models or multi-view image diffusion models to generate 4D content from text prompts \citep{4dfy, stag4d, text-to-4d}. However, these approaches largely focus on generating dynamic objects in blank backgrounds. A notable example is CAT4D \citep{cat4d}, which synthesizes 4D videos conditioned on multiple input modalities using a multi-view video model trained on a curated synthetic multi-view dataset. Similarly, \cite{gcdolly} introduces a framework for novel-view synthesis of dynamic 4D scenes from a single video. This method is trained on synthetic multi-view video data with corresponding camera poses, enabling high-fidelity 4D reconstructions. Concurrently, 4Real~\cite{4real} proposes text-to-4D scene generation pipelines that integrate video diffusion models with canonical 3D Gaussian Splatting (3DGS) \citep{gs3d}, ensuring spatio-temporal consistency in the generated 4D outputs. Furthermore, 4Real-Video~\cite{4real-video} enhance video diffusion models by introducing a parallel camera-temporal token stream and a learnable synchronization layer, which effectively fuses independent tokens to maintain camera and temporal consistency across generated frames. While these \textit{4D video diffusion models} enable camera control and maintain multi-view and temporal consistency, they rely on training a large diffusion model with 4D data, which is limited in availability and costly to obtain (see Table~\ref{tab:model_comparison}).

\section{Zero4D}

\begin{figure*}
    \centering
    \includegraphics[width=1\linewidth]{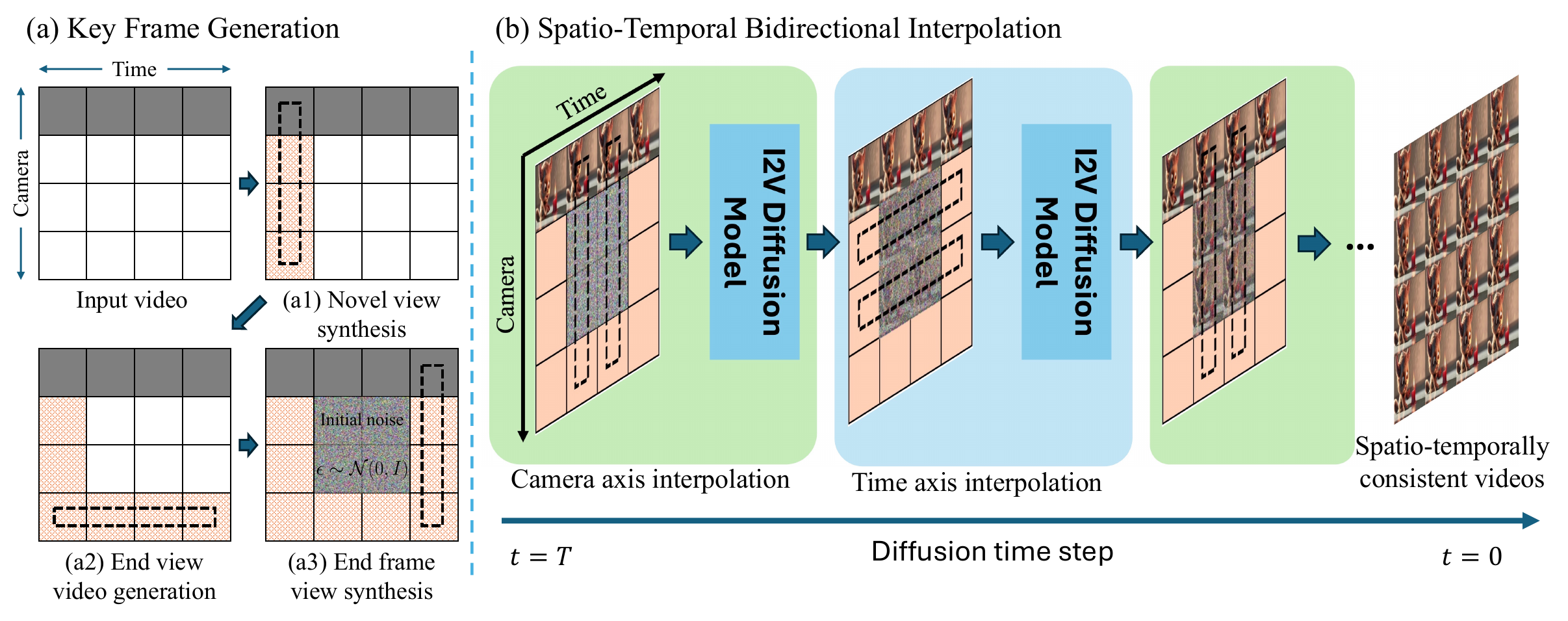}
    \vspace{-20pt}
    \caption{\textbf{Generation pipeline of Zero4D:} \textbf{(a)} \textbf{Key frame generation step:}  Starting from the input video(shown as the gray-shaded row), we sequentially generate boundary frames—novel view synthesis, end-view video generation, and end-frame view synthesis—where each step leverages the results of the previous one.
    \textbf{(b)} \textbf{Spatio-temporal bidirectional interpolation step:} Starting from the noisy frames, we alternately perform camera-axis and time-axis interpolation, each conditioned on boundary frames, to progressively denoise the 4D grid. Through this bidirectional process, noisy latents are refined into globally coherent spatio-temporal videos. The detailed algorithm is described in Algorithm~\ref{alg:zero4d_pipeline}.
    \vspace{-5pt}
    }
    \label{fig:main_pipe}
\end{figure*}

Let $x[i,j] \in \mathbb{R}^{H \times W}, i=1,\cdots, N, j=1,\cdots, F$ denotes
the image at the $i$-th camera viewpoint and the $j$-th temporal frame, where $H$ and $W$ denote the height and width of the image, respectively (see Fig.~\ref{fig:main_pipe}(a)).
Then,  the input video captured from a single camera viewpoint $c$ is denoted as $x[c,:]$,
whereas the multi-view images at the temporal frame $f$ are represented by $x[:,f]$.
The goal of Zero4D is then to populate the spatio-temporal video grid (or camera-time grid) $x[:,:]$ by generating frames across multiple camera poses.
The key innovation is that the spatio-temporal grid can be populated entirely at inference time, without any training—a task once thought impossible.
As illustrated in Fig.~\ref{fig:main_pipe}, the overall reconstruction pipeline of Zero4D is composed
of two steps: 1) key frame generation and 2) spatio-temporal bidirectional interpolation along the time and camera axes in an alternating manner. 
In this section, we describe each in detail.

\subsection{Key frame generation}
As shown in Fig.~\ref{fig:main_pipe}(a), the key frame generation is achieved through three steps. Specifically, given a input video denoted by $x[1,:]$, 
     we first perform novel-view synthesis,  followed by end-view video frame generation.
     These two steps are achieved through diffusion sampling, guided by warped views. 
 Finally, we complete the rightmost column using diffusion-based interpolation sampling.

\noindent\textbf{Novel view synthesis (a1).}
First, we synthesize novel view video \( x[:,1] \) from the first frame image \( x[1,1] \) using the I2V diffusion model. Here, we incorporate the warped frames \( x_w[:,1] \) as guidance to ensure the generated novel views align with the warped images from input video. The warped frames  \( x_w[:,:] \) are computed as follows.
Given an input video \( x[1,:] \), we generate novel views by first estimating a per-frame depth map \( D[1,:] \) using a video depth estimation model~\cite{depthcrafter}, which ensures a shared scale and shift across all frames. This depth information enables depth-based geometric warping, wherein each frame of the input video is unprojected into 3D space and reprojected into a target viewpoint in \( p(n) \in \mathcal{P}_N  \) where
$\mathcal{P}_N$ defines the desired set of camera views. 
This produces the warped frames:
\begin{equation}
    x_w[n,i] = \mathcal{W}\left(x[1,i], D[1,i], p(n), K\right),
    \quad i = 1, \dots, F,
\end{equation}
for $n=1,\cdots, N$,
where $K$ is the intrinsic camera matrix.
The warping function $\mathcal{W}(\cdot)$ unprojects each pixel using its estimated depth and reprojects it into the target view. Formally, for each pixel location $r_i$ in the $i$-view, the warped pixel location $r_j$ in the novel-view at the $j$-th camera location is computed as:
\begin{equation}
    r_j = K P_{i \to j} D_i(r_i) K^{-1} r_i,
\end{equation}
where $P_{i \to j}$ is the transformation from the input to the novel-view, and $D_i(r_i)$ is the depth at $r_i$. Since $r_j$ may not align exactly with integer pixel locations, interpolation is applied to assign pixel values. However, missing regions (e.g., occlusions from depth-based projection) often appear in \( x_w \). 
To address this, we utilize a video diffusion model\citep{svd} parameterized by \( \theta \) to inpaint the missing regions and ensure consistency within the 4D video grid.  This can be considered as conditional sampling under the condition
of the warped image, occlusion mask, and the input video conditioning.
For the case of novel-view synthesis at the temporal frame index $j=1$, this corresponds to
\vspace{-2pt}
\begin{equation}
    x[:,1] \sim p_{\theta}\left(x[:,1] \mid x_w[:,1], m_w[:,1], c[1,1]\right),
\end{equation}
where $p_\theta$ corresponds to the conditional distribution from the trained diffusion model, 
 \( m_w[:,1] \) is an occlusion mask that identifies missing pixels, and \( c[1,1] \) is conditioned embedding vector from $x[1,1]$.
 The specific details of conditional video diffusion sampling will be described in Section~\ref{sec:condition}.

\begin{algorithm}[t]
\small
\setlength{\textfloatsep}{1pt}
\caption{Zero4D overall pipeline}
\label{alg:zero4d_pipeline}
\LinesNumbered
\begin{spacing}{1.2} 

\SetInd{0.7em}{0.7em} 

\begin{minipage}{\linewidth} 

\KwIn{Input video $x[1,:]$, warped views $x_w[:,:]$, masks $m_w[:,:]$, Interpolator $I_{\theta}$}
\KwOut{4D video grid $x_0[:,:] \in \mathbb{R}^{N \times F}$}

\BlankLine
\textbf{Stage A — Boundary/Keyframe generation}\;

$x[:,1] \sim p_{\theta}(x[:,1] \mid x_w[:,1], m_w[:,1], c[1,1])$\tcp*{(a1)}
$x[N,:] \sim p_{\theta}(x[N,:] \mid x_w[N,:], m_w[N,:],c[N,1])$\tcp*{(a2)}
\For(\tcp*[f]{(a3)}){$t \gets T$ \KwTo $0$}{
  $x_{t-1}[:,F] \leftarrow I_{\theta}(x_t[:,F], \sigma_t, c[1,F], c[N,F], x_w[:,F])$
}
$c[:,1],c[N,:],c[:,F] \leftarrow \text{Encode}(\{ x[:,1], x[N,:], x[:,F]\})$\;

\BlankLine
\textbf{Stage B — Spatio–temporal bidirectional interpolation}\;

$x_T[:,:] \sim \mathcal{N}(0,I)$\;

\For{$t \gets T$ \KwTo $1$}{
   \For(\tcp*[f]{Camera-axis interpolation}){$i \gets 1$ \KwTo $F$}{
      \hspace{-0.5em}$x_{t-1}[:,i] $\hspace{-0.5em} $\leftarrow $\hspace{-0.4em} $ I_{\theta}(x_t[:,i], \sigma_t, c[1,i], c[N,i], x_w[:,i], m_w[:,i])$
      \hspace{-0.5em}$x_t[:,i] \leftarrow x_{t-1}[:,i] + \sqrt{\sigma_t^2 - \sigma_{t-1}^2}\,\epsilon$
   }
   \For(\tcp*[f]{Time-axis interpolation}){$j \gets 1$ \KwTo $N$}{
      \hspace{-0.5em}$x_{t-1}[j,:]\hspace{-0.3em} \leftarrow\hspace{-0.3em} I_{\theta}(x_t[j,:], \sigma_t, c[j,1], c[j,F], x_w[j,:], m_w[j,:])$}
}
\Return{$x_0[:,:]$}

\end{minipage}
\end{spacing}
\end{algorithm}

\noindent\textbf{End view video generation (a2).}
Similarly, we can synthesize the end-view video $x[N,:]$ from the generated view $x[N,1]$ utilizing warp-guided diffusion sampling.
\begin{equation}
    x[N,:] \sim p_{\theta}\left(x[N,:] \mid x_w[N,:], m_w[N,:], c[N,1]\right).
\end{equation}
This process follows the same video sampling approach as first-frame novel-view synthesis; however, it differs in that it synthesizes the video from the final camera position. 

\noindent\textbf{End frame novel-view synthesis (a3).}
Finally, we generate video at the end-frame novel-view \( x[:,F] \), which constitutes the rightmost column of the 4D grid in Fig.~\ref{fig:main_pipe}(a). Given that we already have \( x[1,F] \) from the input video and the synthesized end-view frame \( x[N,F] \) derived from \( x[N,:] \), we incorporate both images to enhance consistency. To this end, we repurpose a video interpolation method that simultaneously conditions on both \( c[1,F] \) and \( c[N,F] \) for novel-view synthesis. During interpolation, we further incorporate the warped image and its mask to fully exploit the available prior information. In particular, we synthesize the last column $x[:,F]$ leveraging video diffusion interpolation method~\cite{yang2024vibidsampler}:
\begin{equation}
\begin{aligned}
    x_{t-1}[:,F] &= I_{\theta}\!\left(
        x_t[:,F],\, \sigma_t,\, c[1,F],\, c[N,F],\, x_w[:,F]
    \right) \\
    &\text{for } t = T \to 0 .
\end{aligned}
\end{equation}
where $I_{\theta}$ denotes the one-step denoising using video interpolation. The final novel-view frame $x[:,F]$ is obtained iteratively by applying $I_{\theta}$ over diffusion time steps $t = T \to 0$. The detailed implementation of the interpolation process is provided in Algorithm~2
of Appendix~A.4

\subsection{Spatio-temporal bidirectional interpolation}
As shown in Fig.~\ref{fig:main_pipe}(b), once the keyframes are generated, the remaining task is to fill in the missing sampling
grid at the center so the final resulting 4D video remains consistent across both the camera and time axes.
Accordingly, it is essential to perform conditioned sampling using the key frames and adjacent frames from the camera and temporal axes. However, a naive image-to-video diffusion model can only condition on a single or two end frames. To address this challenge, we first repurpose a video interpolation approach to generate spatio-temporally consistent samples under multi-view conditions. The key idea is to alternate interpolation along both the camera and time axes, thereby guiding the overall diffusion trajectory to satisfy the multiple constraints from the keyframes. In this work, we leverage ViBiDSampler~\citep{yang2025vibidsampler} as the interpolator, with implementation details provided in Appendix~A.4

\noindent{\textbf{Camera axis interpolation.}} Starting from the initial noise $x_T[:,:] \sim \mathcal{N}(0,I)$, 
we select a specific frame in the 4D grid (a column) $x_t[:,i]$, and perform an interpolation denoising process(\ref{eq:camera_interpol_denoise}) using the edge-frame conditions $c[1,i]$ and $c[N,i]$:
\begin{equation}
    x_{t-1}[:,i] \hspace{-3pt}\leftarrow I_{\theta}( x_t[:,i], \sigma_t, c[1,i], c[N,i], x_w[:,i])
    \label{eq:camera_interpol_denoise}
\end{equation}
Here, the image condition $c[1,i]$ is applied first, along with the warped view to guide the diffusion denoising step. The video is then perturbed with noise again, flipped along the camera axis, and subjected to another diffusion denoising step using $c[N,i]$ as the condition. Through these two conditioning steps, $x_t[:,i]$ integrates information from both $c[1,i]$ and $c[N,i]$, enabling interpolation-based denoising that preserves consistency across the camera axis. Before proceeding to time axis interpolation, we apply a re-noising step to ensure smooth transitions across generated frames.
\begin{figure*}[!t]
  \centering
  \begin{minipage}[b]{0.16\linewidth}
  \centering
  {Input video}
  \end{minipage}
  \begin{minipage}[b]{0.31\linewidth}
  \centering
  {Camera orbit controls}
\end{minipage}
\begin{minipage}[b]{0.16\linewidth}
  \centering
  {Input video}
\end{minipage}
\begin{minipage}[b]{0.31\linewidth}
  \centering
  \hspace{1pt}%
  {Dolly and transition controls}
\end{minipage}
  
  \begin{minipage}[b]{0.155\linewidth}
    \centering
    \animategraphics[loop,width=1.08\linewidth,height=0.68\linewidth,keepaspectratio]{8}{Videos/pixar/org/}{00}{24}
  \end{minipage}\hspace{1pt}
  \begin{minipage}[b]{0.155\linewidth}
    \centering
    \animategraphics[loop,width=1.08\linewidth,height=0.68\linewidth,keepaspectratio]{8}{Videos/pixar/left/}{00}{24}
  \end{minipage}\hspace{1pt}
  \begin{minipage}[b]{0.155\linewidth}
    \centering
    \animategraphics[loop,width=1.08\linewidth,height=0.68\linewidth,keepaspectratio]{8}{Videos/pixar/right/}{00}{24}
  \end{minipage}\hspace{1pt}
  \begin{minipage}[b]{0.155\linewidth}
    \centering
    \animategraphics[loop,width=1.08\linewidth,height=0.68\linewidth,keepaspectratio]{8}{Videos/old_couple/org/}{00}{24}
  \end{minipage}\hspace{1pt}
  \begin{minipage}[b]{0.155\linewidth}
    \centering
    \animategraphics[loop,width=1.08\linewidth,height=0.68\linewidth,keepaspectratio]{8}{Videos/old_couple/in/}{00}{24}
  \end{minipage}\hspace{1pt}
  \begin{minipage}[b]{0.155\linewidth}
    \centering
    \animategraphics[loop,width=1.08\linewidth,height=0.68\linewidth,keepaspectratio]{8}{Videos/old_couple/out/}{00}{24}
  \end{minipage}

  \vspace{-0.1cm}

  \begin{minipage}[b]{0.155\linewidth}
    \centering
    \animategraphics[loop,width=1.08\linewidth]{8}{Videos/popcorn/org/}{00}{24}
  \end{minipage}\hspace{1pt}
  \begin{minipage}[b]{0.155\linewidth}
    \centering
    \animategraphics[loop,width=1.08\linewidth]{8}{Videos/popcorn/left/}{00}{24}
  \end{minipage}\hspace{1pt}
  \begin{minipage}[b]{0.155\linewidth}
    \centering
    \animategraphics[loop,width=1.08\linewidth]{8}{Videos/popcorn/right/}{00}{24}
  \end{minipage}\hspace{1pt}
  \begin{minipage}[b]{0.155\linewidth}
    \centering
    \animategraphics[loop,width=1.08\linewidth]{8}{Videos/surf/org/}{00}{24}
  \end{minipage}\hspace{1pt}
  \begin{minipage}[b]{0.155\linewidth}
    \centering
    \animategraphics[loop,width=1.08\linewidth]{8}{Videos/surf/down/}{00}{24}
  \end{minipage}\hspace{1pt}
  \begin{minipage}[b]{0.155\linewidth}
    \centering
    \animategraphics[loop,width=1.08\linewidth]{8}{Videos/surf/up/}{00}{24}
  \end{minipage}
  
  \vspace{0.03cm}

  \begin{minipage}[b]{0.155\linewidth}
    \centering
    \includegraphics[width=1.08\linewidth,height=0.68\linewidth,keepaspectratio]{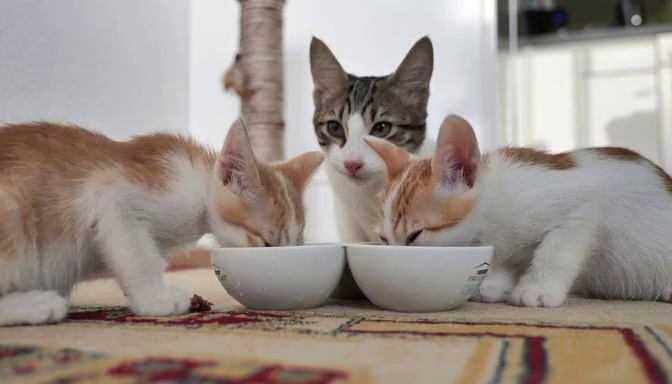}
  \end{minipage}\hspace{1pt}
  \begin{minipage}[b]{0.155\linewidth}
    \centering
    \includegraphics[width=1.08\linewidth,height=0.68\linewidth,keepaspectratio]{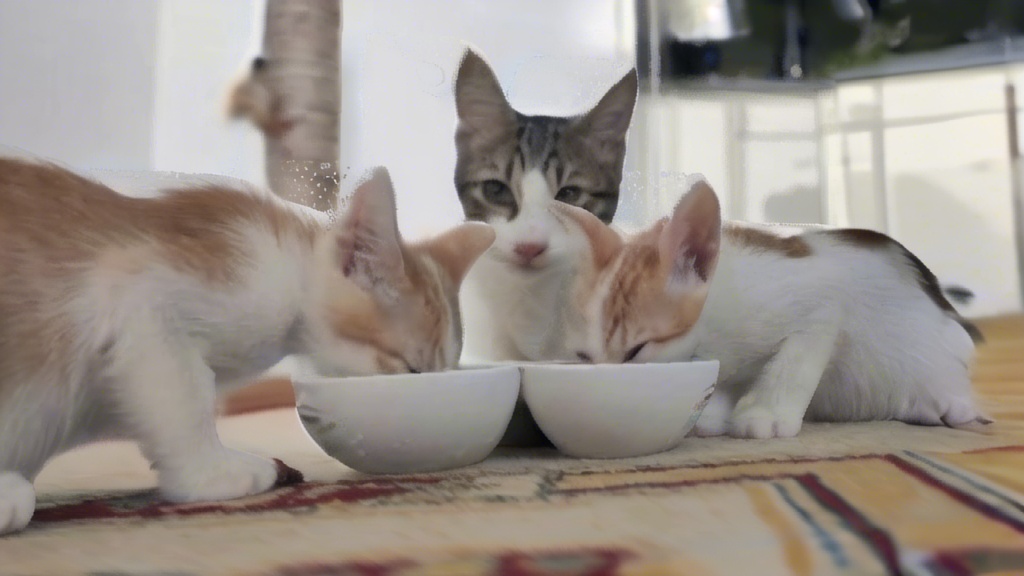}
  \end{minipage}\hspace{1pt}
  \begin{minipage}[b]{0.155\linewidth}
    \centering
    \includegraphics[width=1.08\linewidth,height=0.68\linewidth,keepaspectratio]{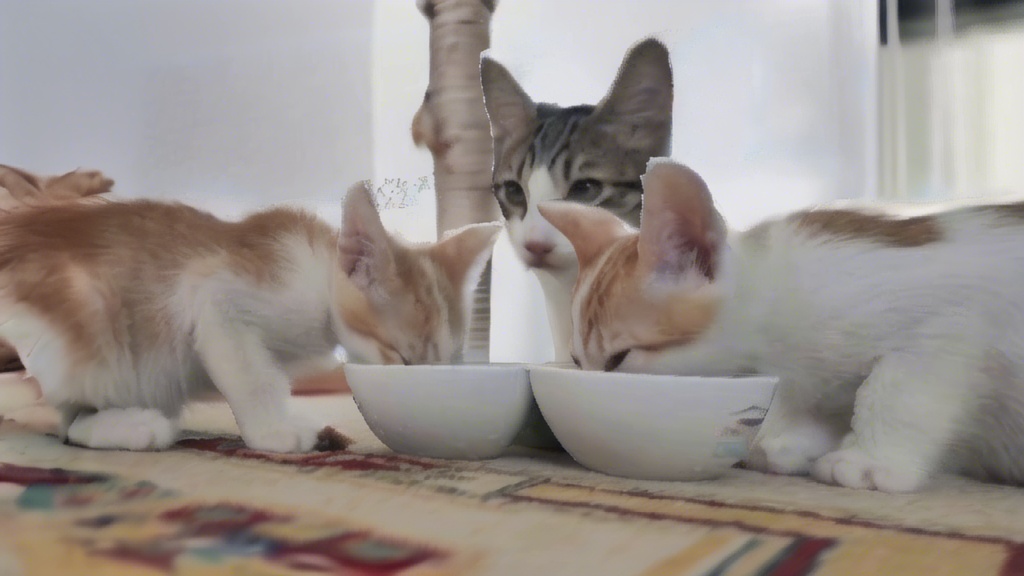}
  \end{minipage}\hspace{1pt}
  \begin{minipage}[b]{0.155\linewidth}
    \centering
    \includegraphics[width=1.08\linewidth,height=0.68\linewidth,keepaspectratio]{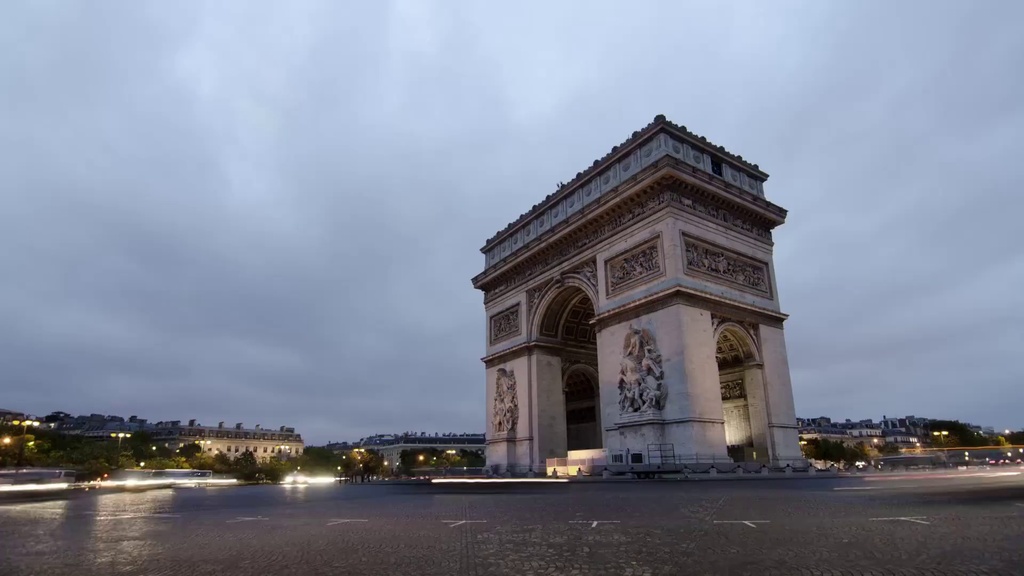}
  \end{minipage}\hspace{1pt}
  \begin{minipage}[b]{0.155\linewidth}
    \centering
    \includegraphics[width=1.08\linewidth,height=0.68\linewidth,keepaspectratio]{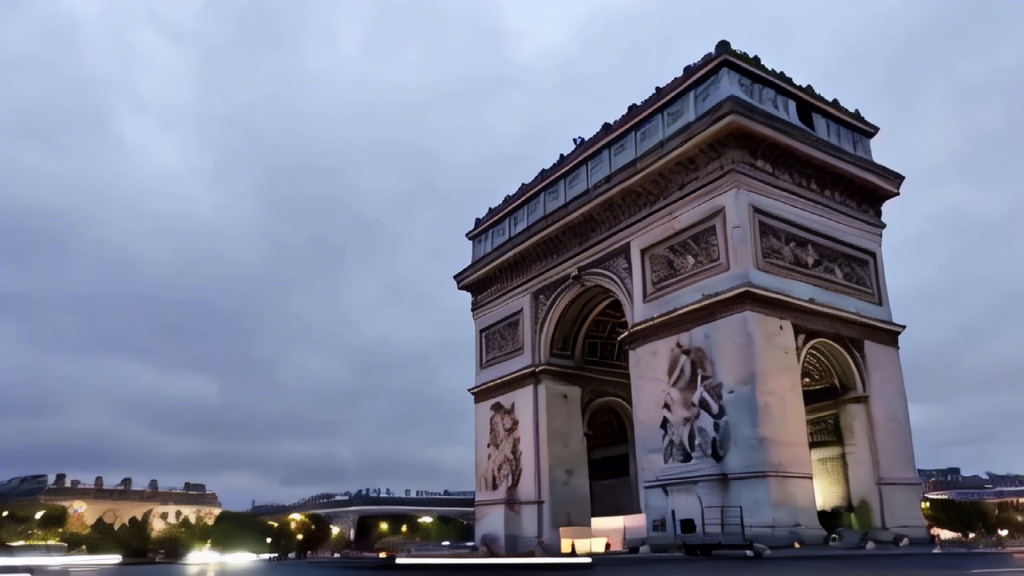}
  \end{minipage}\hspace{1pt}
  \begin{minipage}[b]{0.155\linewidth}
    \centering
    \includegraphics[width=1.08\linewidth,height=0.68\linewidth,keepaspectratio]{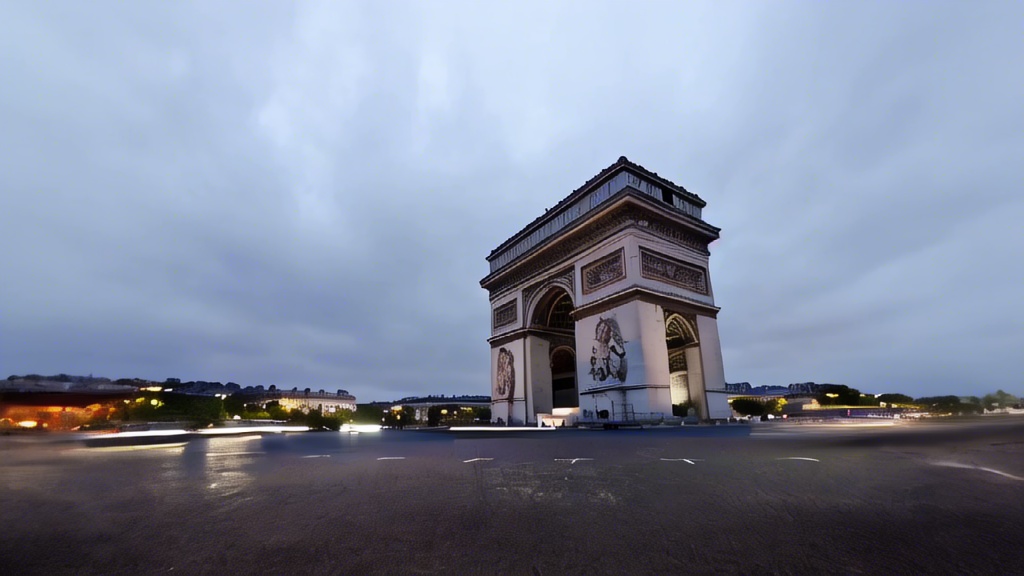}
  \end{minipage}

  \vspace{-0.1cm}

  \begin{minipage}[b]{0.155\linewidth}
    \centering
    \includegraphics[width=1.08\linewidth,height=0.68\linewidth,keepaspectratio]{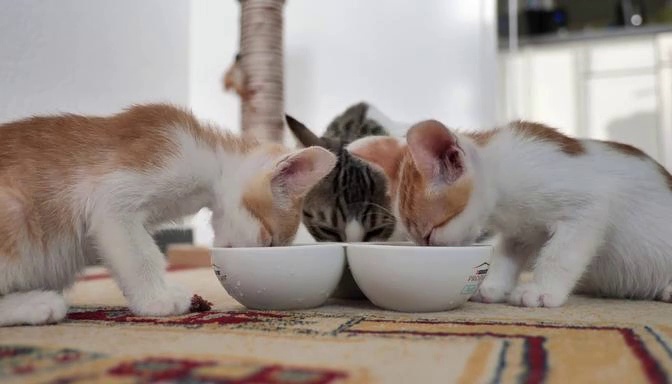}
  \end{minipage}\hspace{1pt}
  \begin{minipage}[b]{0.155\linewidth}
    \centering
    \includegraphics[width=1.08\linewidth,height=0.68\linewidth,keepaspectratio]{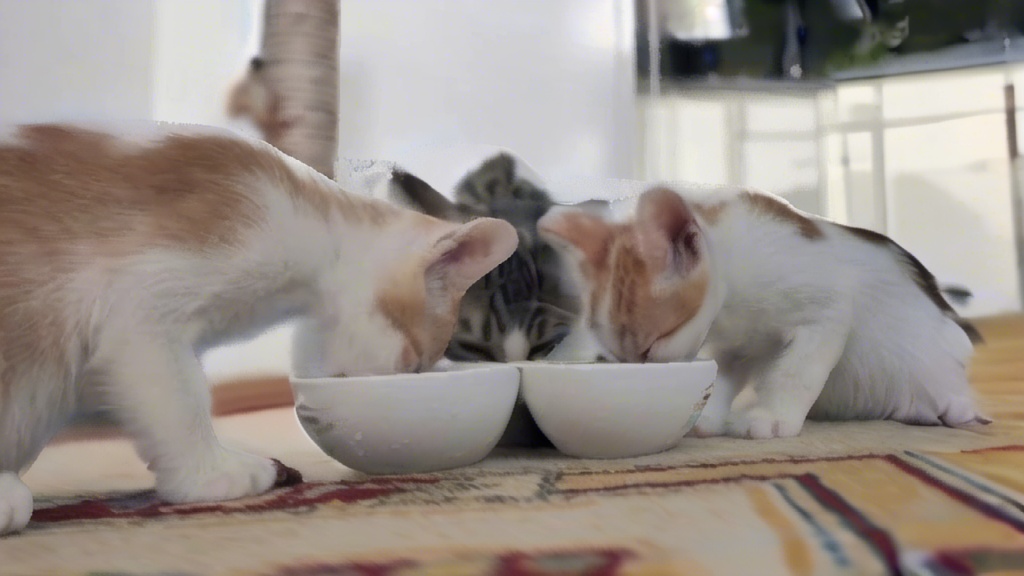}
  \end{minipage}\hspace{1pt}
  \begin{minipage}[b]{0.155\linewidth}
    \centering
    \includegraphics[width=1.08\linewidth,height=0.68\linewidth,keepaspectratio]{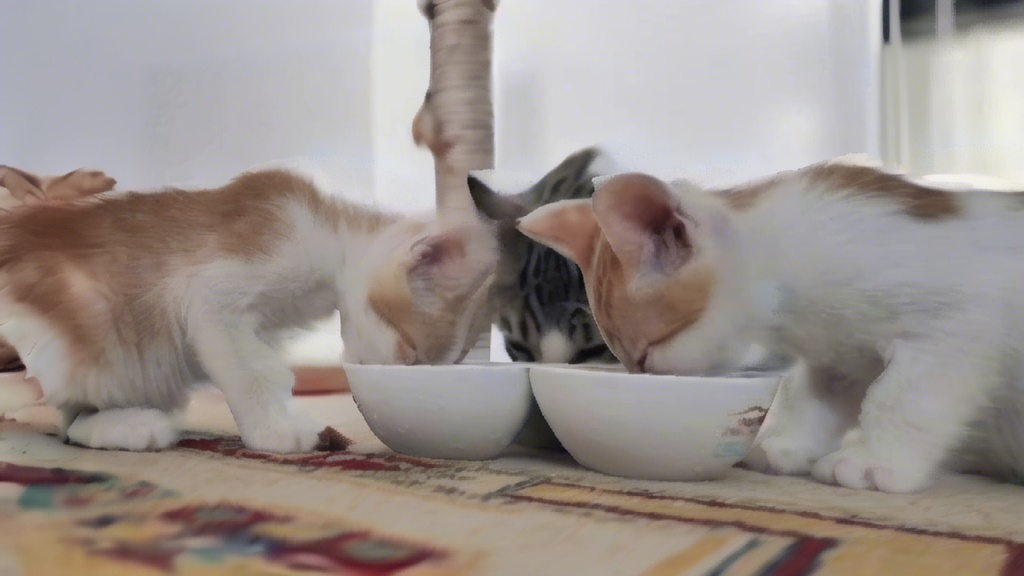}
  \end{minipage}\hspace{1pt}
  \begin{minipage}[b]{0.155\linewidth}
    \centering
    \includegraphics[width=1.08\linewidth,height=0.68\linewidth,keepaspectratio]{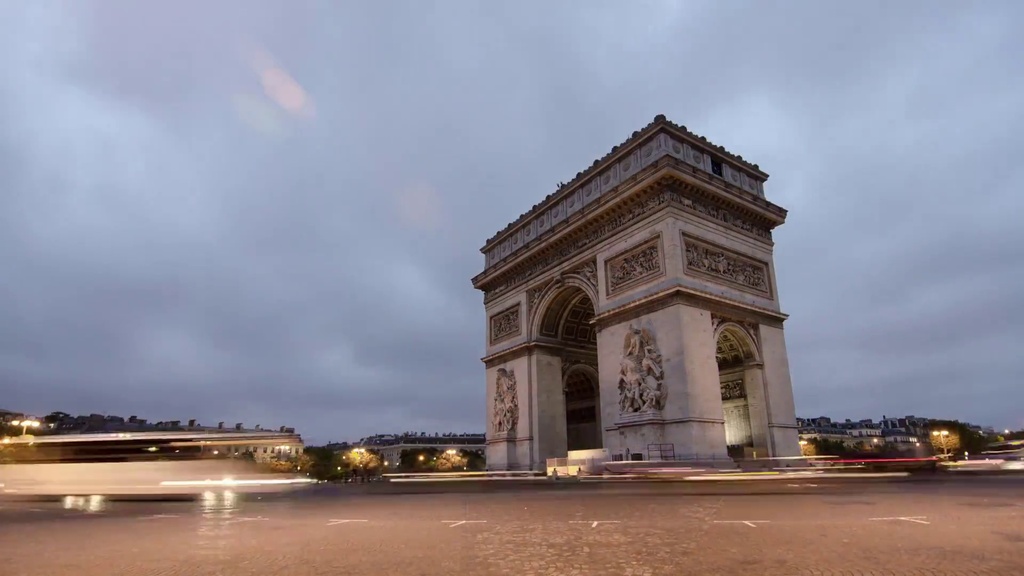}
  \end{minipage}\hspace{1pt}
  \begin{minipage}[b]{0.155\linewidth}
    \centering
    \includegraphics[width=1.08\linewidth,height=0.68\linewidth,keepaspectratio]{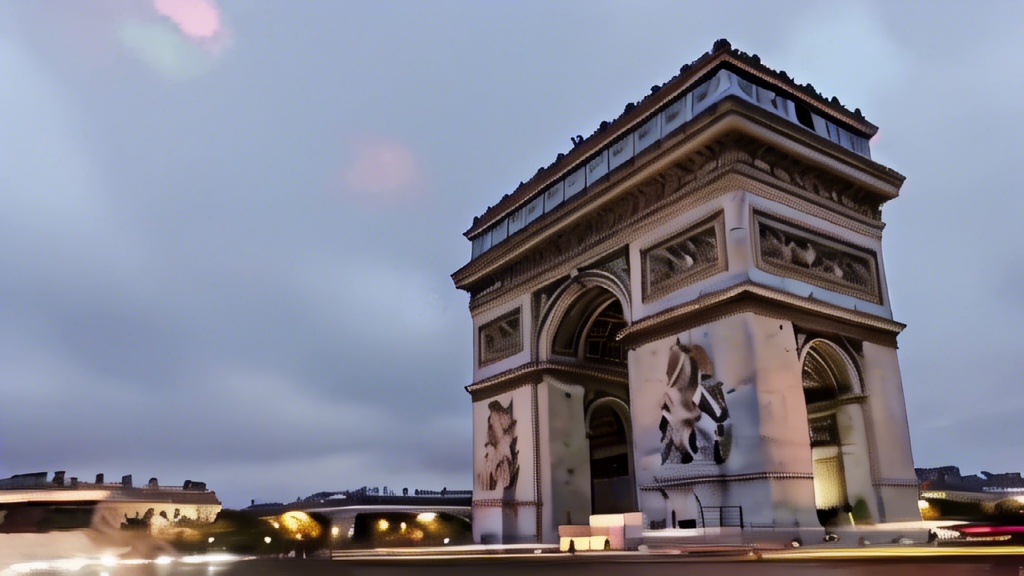}
  \end{minipage}\hspace{1pt}
  \begin{minipage}[b]{0.155\linewidth}
    \centering
    \includegraphics[width=1.08\linewidth,height=0.68\linewidth,keepaspectratio]{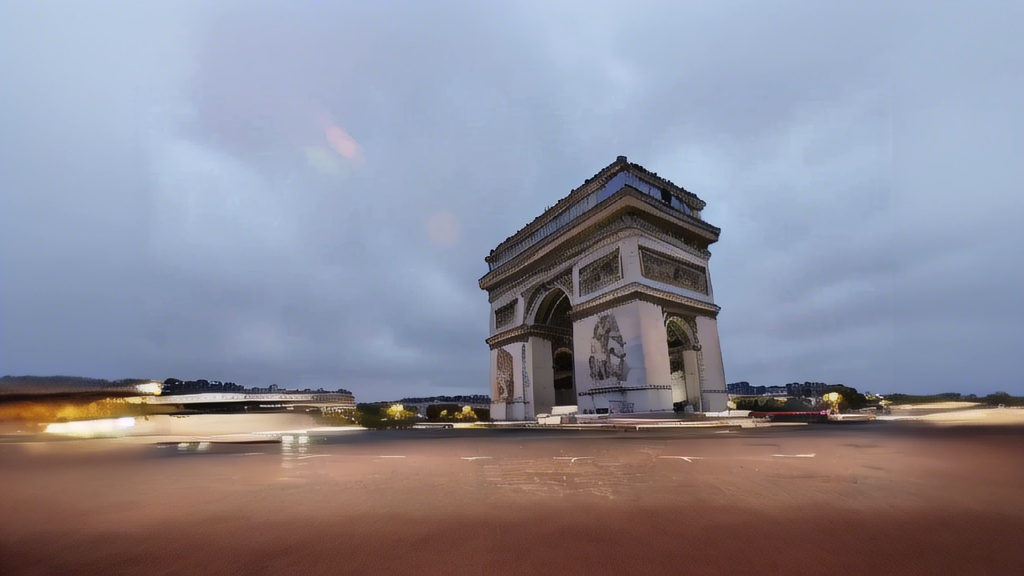}
  \end{minipage}

  \vspace{-0.1cm}

  \begin{minipage}[b]{0.155\linewidth}
    \centering
    \includegraphics[width=1.08\linewidth,height=0.68\linewidth,keepaspectratio]{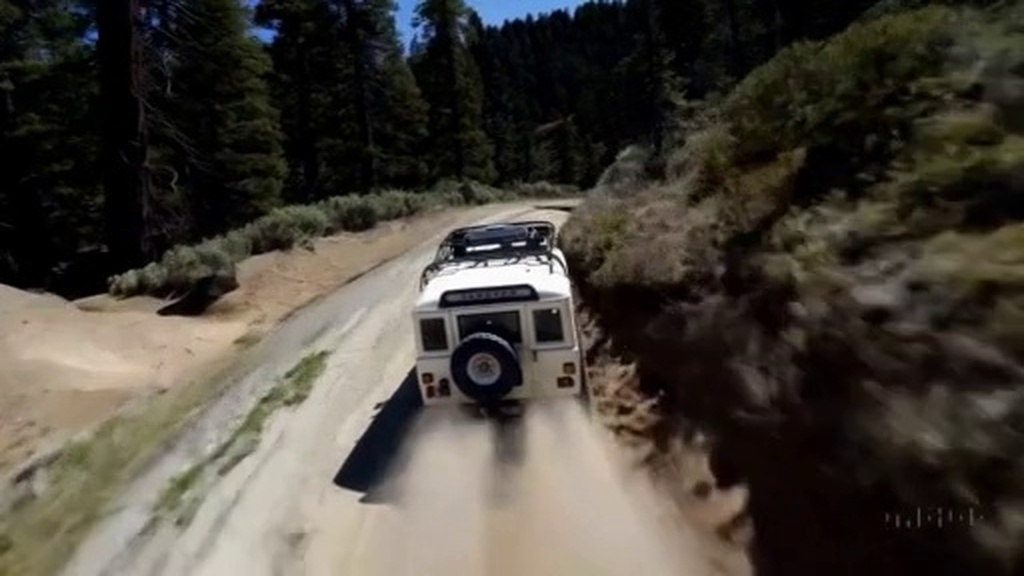}
  \end{minipage}\hspace{1pt}
  \begin{minipage}[b]{0.155\linewidth}
    \centering
    \includegraphics[width=1.08\linewidth,height=0.68\linewidth,keepaspectratio]{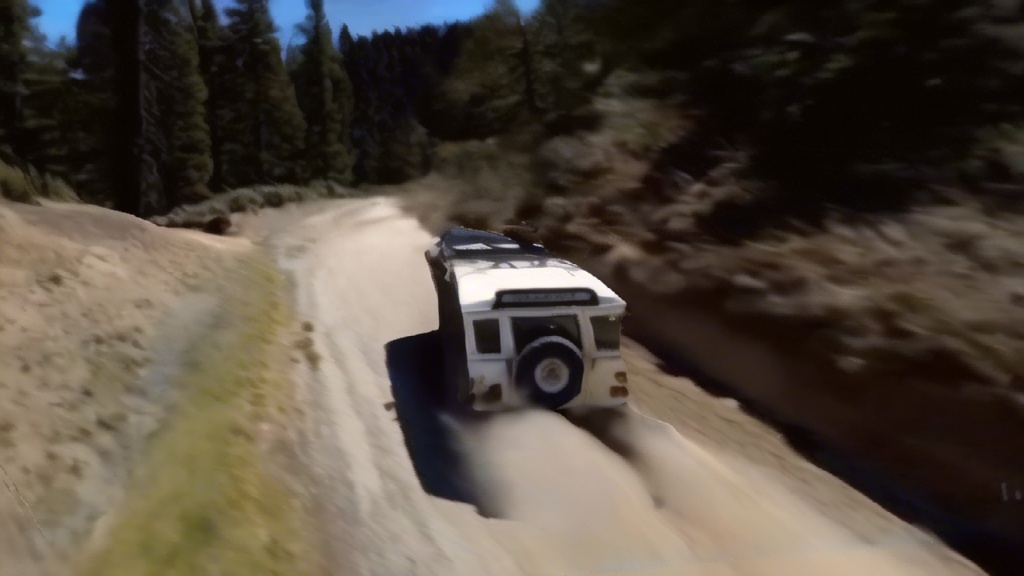}
  \end{minipage}\hspace{1pt}
  \begin{minipage}[b]{0.155\linewidth}
    \centering
    \includegraphics[width=1.08\linewidth,height=0.68\linewidth,keepaspectratio]{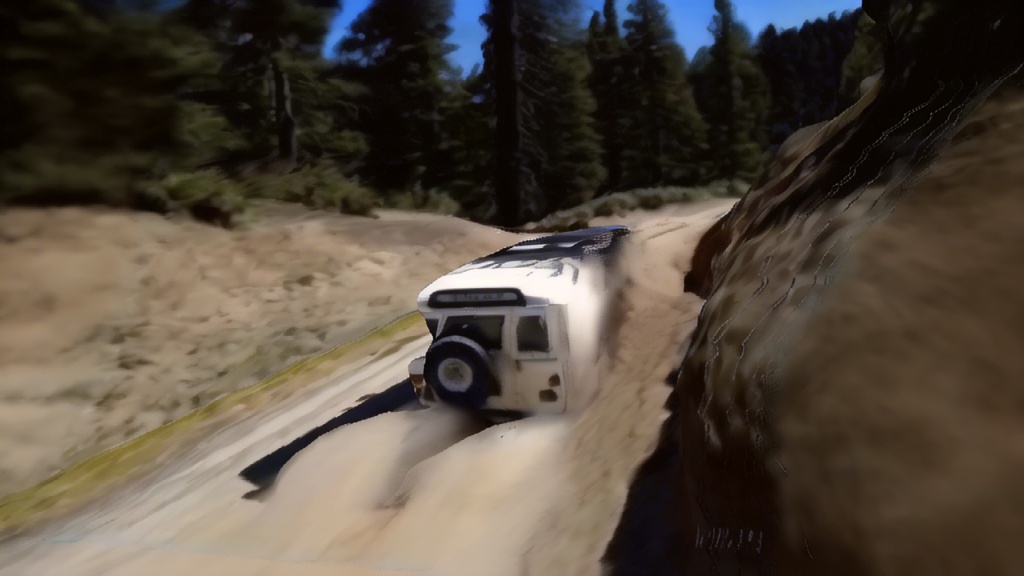}
  \end{minipage}\hspace{1pt}
  \begin{minipage}[b]{0.155\linewidth}
    \centering
    \includegraphics[width=1.08\linewidth,height=0.68\linewidth,keepaspectratio]{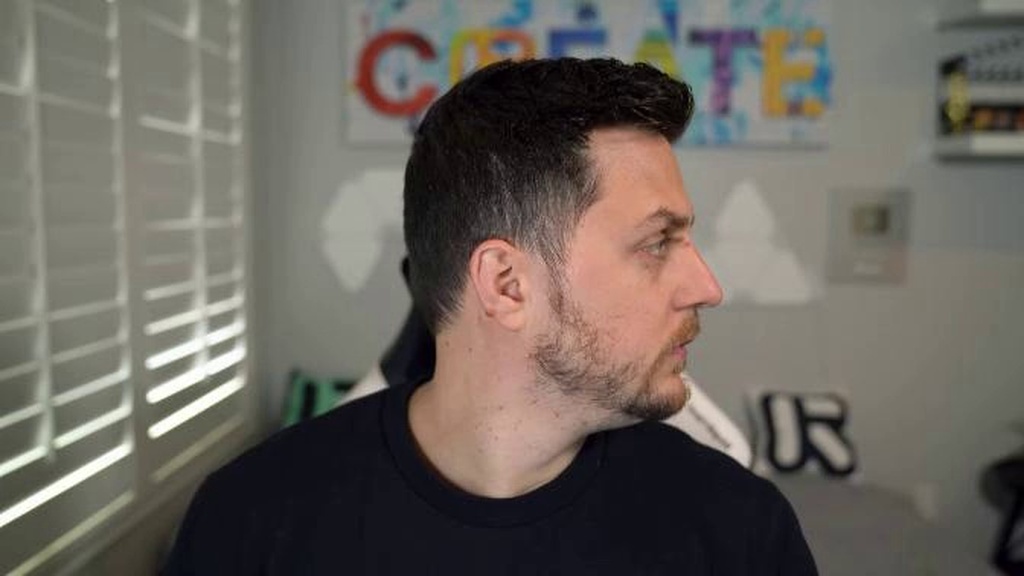}
  \end{minipage}\hspace{1pt}
  \begin{minipage}[b]{0.155\linewidth}
    \centering
    \includegraphics[width=1.08\linewidth,height=0.68\linewidth,keepaspectratio]{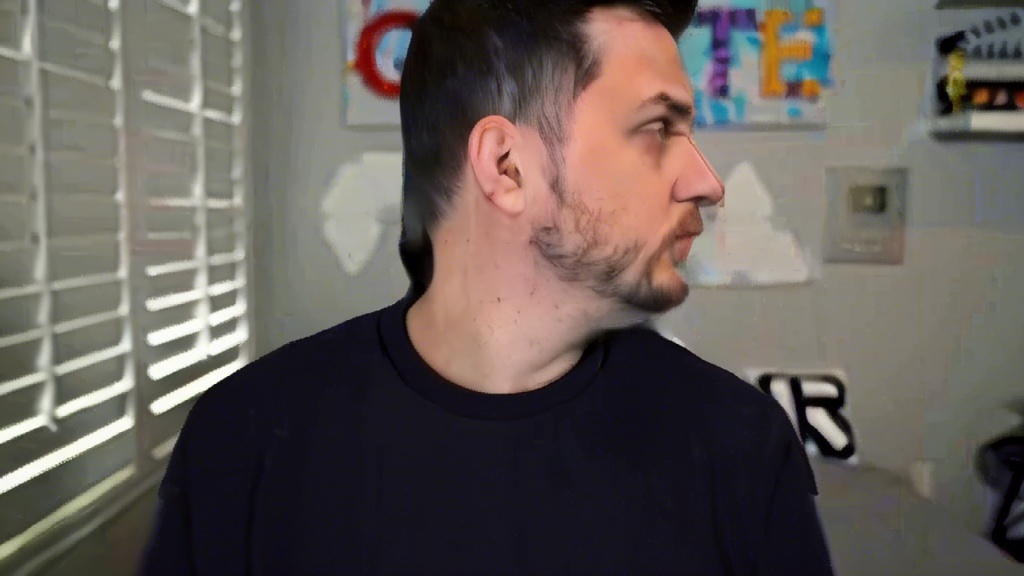}
  \end{minipage}\hspace{1pt}
  \begin{minipage}[b]{0.155\linewidth}
    \centering
    \includegraphics[width=1.08\linewidth,height=0.68\linewidth,keepaspectratio]{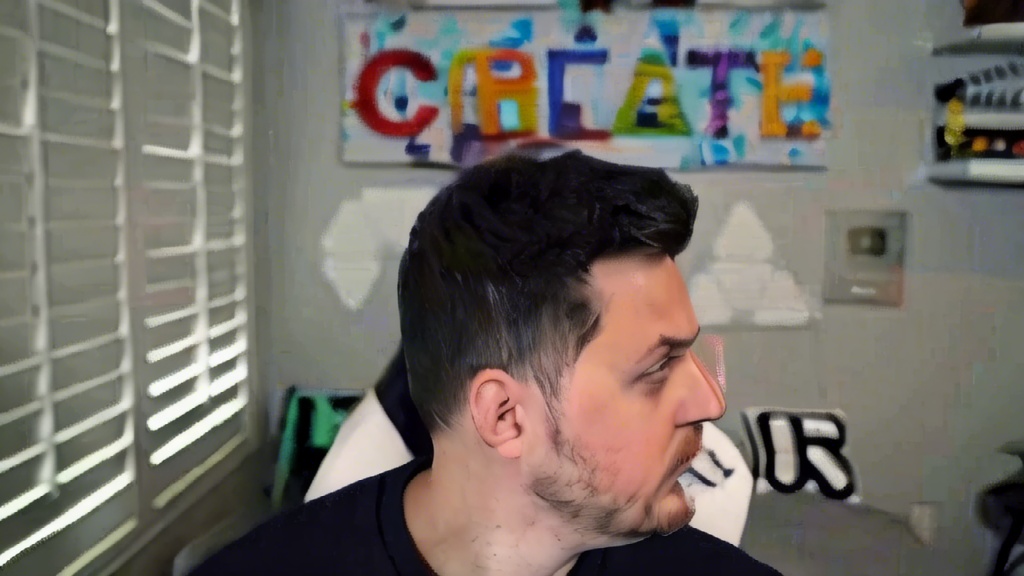}
  \end{minipage}

  \vspace{-0.1cm}

  \begin{minipage}[b]{0.155\linewidth}
    \centering
    \includegraphics[width=1.08\linewidth,height=0.68\linewidth,keepaspectratio]{Images/suv_org_2.jpg}
  \end{minipage}\hspace{1pt}
  \begin{minipage}[b]{0.155\linewidth}
    \centering
    \includegraphics[width=1.08\linewidth,height=0.68\linewidth,keepaspectratio]{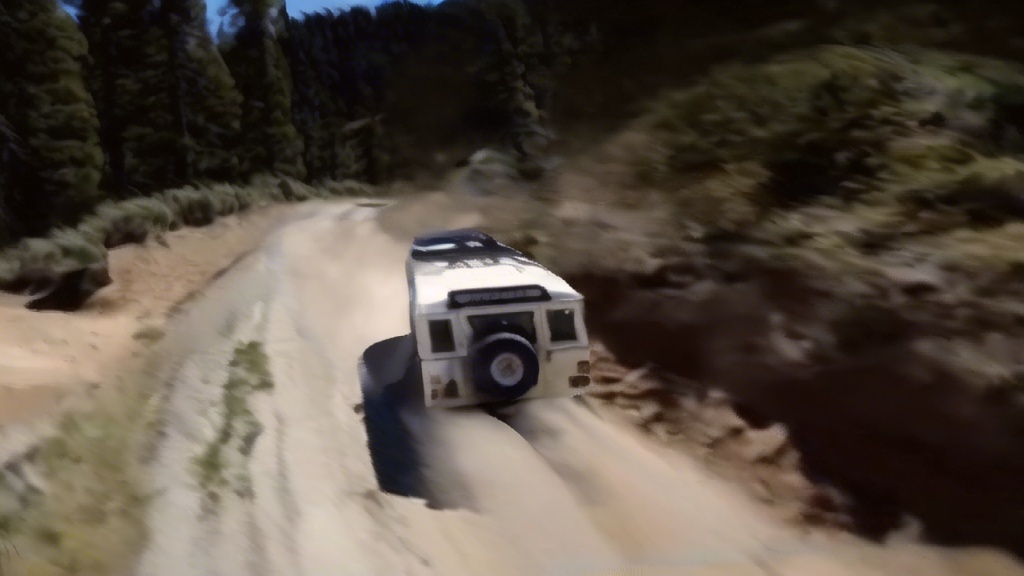}
  \end{minipage}\hspace{1pt}
  \begin{minipage}[b]{0.155\linewidth}
    \centering
    \includegraphics[width=1.08\linewidth,height=0.68\linewidth,keepaspectratio]{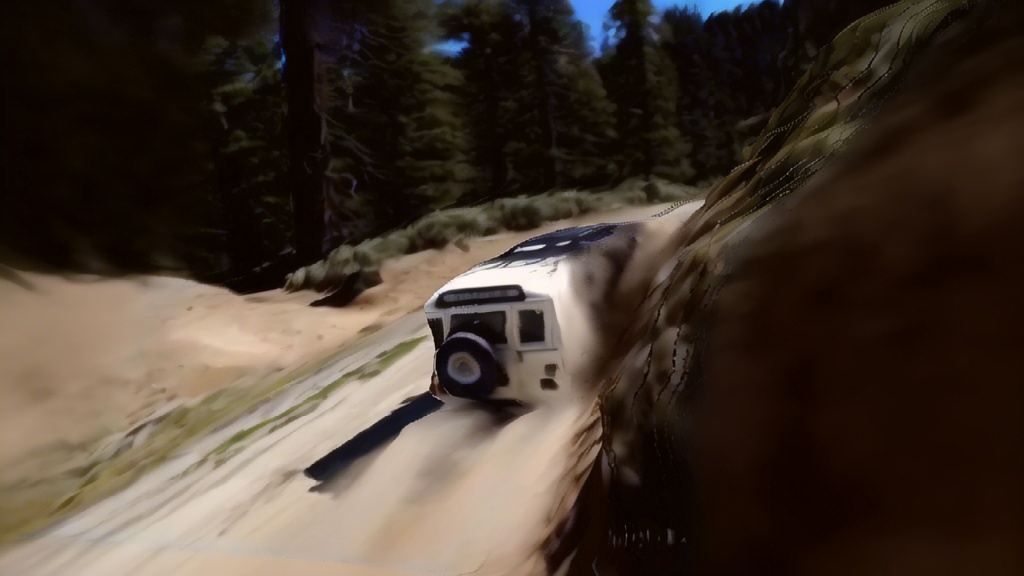}
  \end{minipage}\hspace{1pt}
  \begin{minipage}[b]{0.155\linewidth}
    \centering
    \includegraphics[width=1.08\linewidth,height=0.68\linewidth,keepaspectratio]{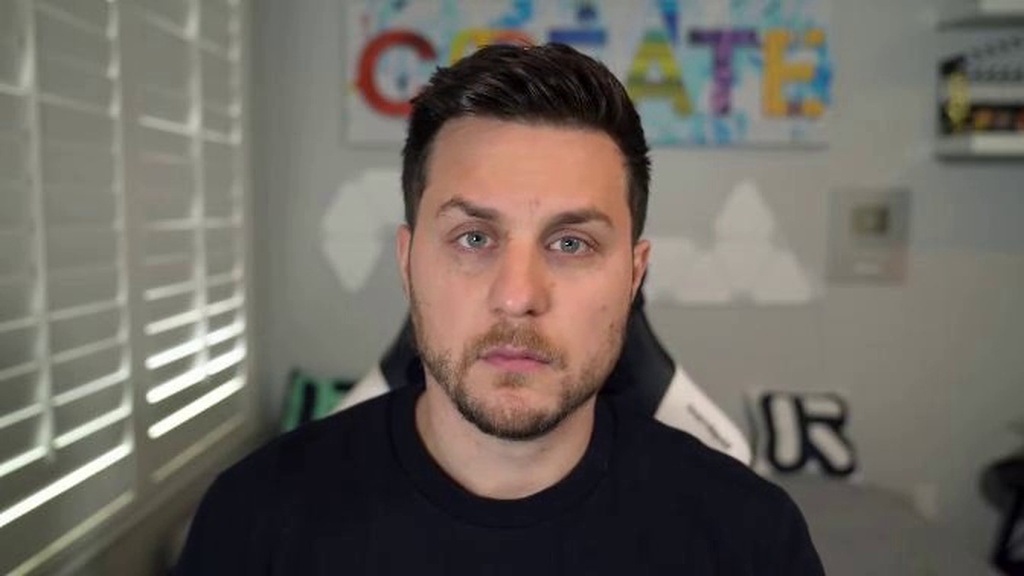}
  \end{minipage}\hspace{1pt}
  \begin{minipage}[b]{0.155\linewidth}
    \centering
    \includegraphics[width=1.08\linewidth,height=0.68\linewidth,keepaspectratio]{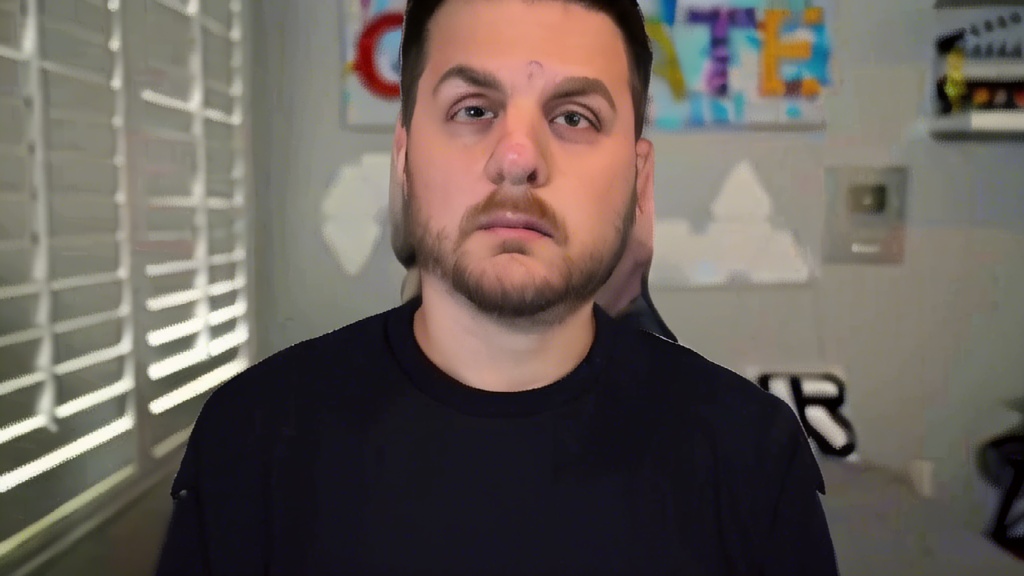}
  \end{minipage}\hspace{1pt}
  \begin{minipage}[b]{0.155\linewidth}
    \centering
    \includegraphics[width=1.08\linewidth,height=0.68\linewidth,keepaspectratio]{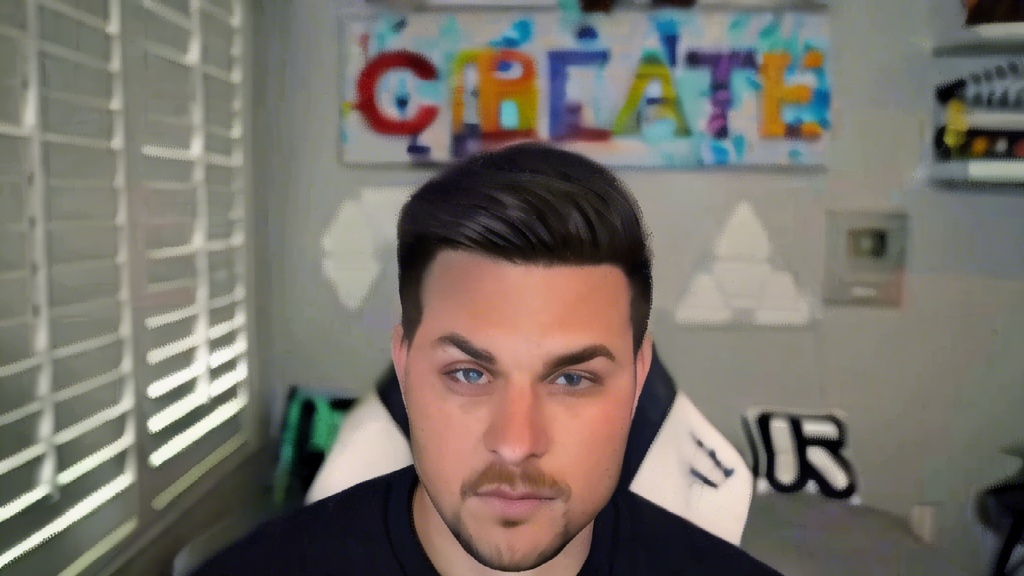}
  \end{minipage}
  
  \caption{\textbf{Result from Zero4D.} 
  Our model generates high-quality 4D videos from a single input video, enabling diverse camera motions such as orbit, transition, and dolly movements. As illustrated, the synthesized videos maintain spatial and temporal consistency across multiple views and frames, effectively rendering novel perspectives that are not present in the original input.  \textit{Best viewed with Acrobat Reader. Click first two rows' images to play the video clip.}
  }
  \label{fig:videos_and_images}
\end{figure*}

\noindent\textbf{Time axis interpolation.}
After ensuring spatial consistency across the camera axis, we interpolate frames along the time axis to maintain temporal coherence. For each row $x_t[j, :]$ in the 4D grid, we perform an interpolation denoising (\ref{eq:time_interpol_denoise}) using the start and end frame conditions $c[j, 1]$ and $c[j, F]$. \begin{equation}
    x_{t-1}[j, :] \hspace{-3pt}\leftarrow I_{\theta} \bigl( x_t[j, :], \sigma_t, c[j, 1], c[j, F], x_w[j, :] \bigr)
    \label{eq:time_interpol_denoise}
\end{equation}
Initially, $c[j, 1]$ is applied along with the warped view to guide the diffusion denoising step. The frame is then perturbed with noise, flipped along the time axis, and another diffusion denoising step is performed using $c[j, F]$ as the condition. Through this bidirectional conditioning process, $x_t[j, :]$ effectively integrates information from both $c[j, 1]$ and $c[j, F]$, facilitating interpolation-based denoising that ensures smooth transitions along the time axis. 
Throughout the  diffusion steps, we perform denoising by alternating interpolation along the camera axis and time axis. This approach maintains global coherence while ensuring consistency in multi-view video generation.  

\subsection{Details of conditional video diffusion}
\label{sec:condition}
Our work is built upon Stable Video Diffusion (SVD)~\citep{svd}, an image-to-video diffusion model that follows the principles of the EDM framework~\citep{edm}. 
SVD utilizes an iterative denoising approach based on an Euler step method, which progressively transforms a Gaussian noise sample \( x_T \) into a clean signal \( x_0 \):
\begin{equation}
    x_{t-1}(x_t; \sigma_t, c) 
    := \hat{x}_c(x_t) 
       + \frac{\sigma_{t-1}}{\sigma_t}\,\bigl(x_t - \hat{x}_c(x_t)\bigr),
    \label{reverse_process}
\end{equation}
where the  initial noise is $x_T\sim \mathcal{N}(0,I)$,  $\hat{x}_c(x_t)$ is the denoised estimate by Tweedie's formula using the 
score function trained by the neural network parameterized by $\theta$, and $\sigma_t$ is the discretized noise level for each timestep $t \in [0, T]$.

Now, we describe how to modify SVD to enable conditional sampling under the condition on warped image $x_w$,  occlusion mask $m$, and conditioning input $c$.
For convenience, we refer to \( x_t[:,:] \) as \( x_t \). From the formulation of the reverse diffusion sampling process in \eqref{reverse_process}, the reverse diffusion process can be modulated by conditioning on a known scene-prior $x_{\text{known}}$ \citep{repaint}:
\begin{equation}
    \bar{x}_{c}(x_t) = \hat{x}_{c}(x_t) \cdot {m} + {x_{\text{known}}} \cdot (1-m),
\end{equation}
where $m$ is a mask that determines which parts of the scene are known, guiding the denoising process by preserving the warped pixels while allowing the diffusion model to inpaint the missing areas.
In our approach, rather than relying on an externally defined scene-prior $x_{\text{known}}$, we leverage the warped frames $x_w$ obtained from depth-based warping as the conditional guidance. Specifically, we redefine the denoising process by replacing $x_{\text{known}}$ with $x_w$ and substituting $m$ with the occlusion mask $m_w$:
\begin{equation}
    \bar{x}_{c}(x_t) = \hat{x}_{c}(x_t) \cdot {m_w} + {x_w} \cdot (1-m_w).
    \label{eq:warp_guidance}
\end{equation}
Here, the occlusion mask $m_w$ ensures that the visible regions in $x_w$ directly guide the denoising process, while the unseen parts are inpainted using the learned prior.
By incorporating this modified formulation into the reverse diffusion process, we obtain the following sampling update:
\begin{equation}
    x_{t-1}(x_t; \sigma_t, c) 
    \gets \bar{x}_{c}(x_t) 
    + \frac{\sigma_{t-1}}{\sigma_t}\,\bigl(x_t - \hat{x}_c(x_t)\bigr),
\end{equation}
where the target camera viewpoints influence the generated frames through the depth-warped observations \( x_w \), ensuring geometric consistency during video synthesis. 
Throughout the reverse sampling, we iteratively apply this procedure. Additionally, following the approach of \citep{repaint, ext_nvs}, we incorporate resampling annealing to further enhance output quality.

\begin{table*}[t]
    \centering
    \caption{\textbf{Quantitative result in novel view video generation.} We evaluate our method against baselines on VBench, comparing multi-view video results based on novel-view generation from a fixed camera view. Our method achieves the best performance in both frame consistency across videos and image quality of individual frames. (* denotes baselines evaluated with bullet-time re-alignment)}
    \resizebox{1.0\linewidth}{!}{
    \begin{tabular}{l|c c c c c c c}
        \toprule
        \textbf{Method} & 
        \makecell[c]{Subject\\Consistency} $\uparrow$ & 
        \makecell[c]{Background\\Consistency} $\uparrow$ & 
        \makecell[c]{Temporal\\Flickering} $\uparrow$ & 
        \makecell[c]{Motion\\Smoothness} $\uparrow$ &
        \makecell[c]{Dynamic\\Degree} $\downarrow$& 
        \makecell[c]{Image\\Quality} $\uparrow$ & 
        \makecell[c]{Aesthetic\\Quality} $\uparrow$ \\
        \midrule
        SV4D~\cite{sv4d}& 88.76\%& 91.36\%& 94.21\%& 95.28\%& 49.20\%& 46.89\%& 34.36\%\\
        GCD~\cite{gcd}& 90.31\%& 94.13\%& 96.14\%& 93.21\%& \best{19.23\%}& 45.77\%& 32.98\%\\
        TrajectoryAttention~\cite{trajattn}& 88.83\% & 91.42\% & 96.86\% & 97.89\% & 59.50\%& 42.98\% & 37.92\%\\
        TrajectoryCrafter~\cite{trajcrafter}& \second{93.47\%} & \best{96.93\%} & \best{98.42\%} & \best{99.26\%} & \second{21.00\%}& \best{52.10\%} & \best{44.41\%}\\
        Ours& \best{95.55\%} & \second{95.75\%} & \second{97.48\%}& \second{98.34\%}& 27.50\% & \second{51.12\%}& \second{38.22\%}\\
        \midrule
        CameraCtrl*~\cite{cameractrl}& 91.71\% & 91.05\% & 89.98\% & 91.03\% & 98.00\%& 40.12\% & 35.86\%\\
        TrajectoryAttention*~\cite{trajattn}& \second{94.72\%} & \second{94.93\%} & \best{97.61\%} & \second{98.28\%} & \best{27.50\%}& 47.75\% & \best{42.88\%}\\
        TrajectoryCrafter*~\cite{trajcrafter}& 94.71\% & 94.48\% & 94.74\% & 96.81\% & \second{32.50\%}& \second{48.81\%} & 35.86\%\\
        Ours& \best{95.55\%} & \best{95.75\%} & \second{97.48\%} & \best{98.34\%} & \best{27.50\%}& \best{51.12\%} & \second{38.22\%}\\
        \bottomrule
    \end{tabular}
    }
    \label{tab:quant_comparison}
\end{table*}

\vspace{-2pt}

\begin{table*}[t]
    \centering
    \caption{\textbf{Quantitative ablation.} Ablation studies on generated videos show that incorporating all components yields the best performance.}
    \resizebox{\linewidth}{!}{
    \begin{tabular}{l|c c c c c c c c c c}
        \toprule
        \textbf{Method} & ATE (m,$\downarrow$) & RPE-T (m, $\downarrow$) & RPE-R (deg, $\downarrow$) & 
        \makecell[c]{Subject\\Consistency} $\uparrow$ & 
        \makecell[c]{Background\\Consistency} $\uparrow$ & 
        \makecell[c]{Temporal\\Flickering} $\uparrow$ & 
        \makecell[c]{Motion\\Smoothness} $\uparrow$ &
        \makecell[c]{Dynamic\\Degree} $\downarrow$ & 
        \makecell[c]{Image\\Quality} $\uparrow$ & 
        \makecell[c]{Aesthetic\\Quality} $\uparrow$ \\
        \midrule
        Ours & \second{0.190} & \best{0.142} & \second{0.53} & \best{95.55\%} & \best{95.75\%} & \best{97.48\%} & \best{98.34\%} & \best{27.50\%}& \second{51.12\%} & \second{38.22\%}\\
        w/o STBI & \best{0.175} & \second{0.149} & \best{0.34} & \second{93.23\%} & \second{92.63\%} & \second{93.28\%} & \second{95.24\%} & 100\%& \best{52.38\%} & \best{43.21\%}\\
        w/o warp & 0.501 & 0.251 & 0.89 & 93.73\% & 93.38\% & 93.98\% & 96.12\% & \second{47.29\%}& 43.79\% & 36.11\%\\
        \bottomrule
    \end{tabular}
    }
    \label{tab:quant_ablation2}
\end{table*}

\section{Experiments}

We used the SVD \citep{svd} as an I2V model without additional training. The image resolution was fixed at 576×1024, with 25 cameras and a sequence length of 25 frames, a total of multi-view video frames are 625=$25^2$. All frames were generated to form a multi-view video following the target camera trajectory. For depth-based warping, we utilized off-the-shelf depth models ~\cite{depthcrafter} with various camera movements, including orbit controls (right, left), dolly in/out, and vertical transitions (up, down), with further details on the camera movements provided in Appendix~A.3.

\noindent\textbf{Baseline models.}
We compare against state-of-the-art video generation models that support either camera control or multi-view generation:  
(1) CameraCtrl~\citep{cameractrl} is a camera-controllable video diffusion model. Given a single input image, it can synthesize bullet-time videos by following a predefined camera trajectory. (2) TrajectoryCrafter~\citep{trajcrafter}, a representative baseline, synthesizes novel-view and bullet-time videos from warped frames aligned to a target trajectory. (3) TrajectoryAttention~\citep{trajattn} leverages warped video frames from the input video to generate both novel-view and bullet-time videos. (4) SV4D~\citep{sv4d} is an image-to-video diffusion model capable of generating multiple novel-view videos from a single input video. (5) GCD~\citep{gcd} also takes a single video as input and generates novel views of dynamic 4D scenes by controlling azimuth and elevation angles. Several related works~\cite{4real, 4real-video, cat4d, dimensionx} provide no full implementation code, so they are omitted from direct comparison.

\noindent\textbf{Evaluation protocol.}
We evaluate our method in two categories: (1) fixed novel-view video generation and (2) bullet-time video generation. For novel-view evaluation, we adopt VBench~\citep{vbench}, which measures seven aspects of video quality, including identity retention, motion coherence, and temporal consistency. For bullet-time evaluation, we assess 3D consistency using pose errors (ATE, RPE-T, RPE-R)~\citep{pose_error} obtained via COLMAP~\citep{colmap} and MEt3R~\citep{met3r}, a recent metric based on DUSt3R~\citep{DUSt3R} that quantifies geometric consistency from unposed frames. We conducted all experiments on 50 videos randomly sampled from Webvid-10M~\citep{webvid}, comparing ours with baseline models. We also conduct a user study, presented in Appendix~A.1, which shows that our method achieves superior human evaluation scores compared to the baselines.

\subsection{Fixed novel-view video generation}
We evaluate our method in two settings: (1) novel-view generation for video quality, and (2) spatio-temporal consistency for coherence across views and time.

\noindent\textbf{Evaluation of direct novel-view generation.} 
We assess the quality of novel-view videos from fixed target viewpoints using VBench~\citep{vbench}. 
Zero4D retrieves $x[n,:]$ corresponding to a target camera viewpoint $p(n)$ from the 4D video grid $x[:,:]$ synthesized from the input video $x[1,:]$, 
while baselines directly generate $x[n,:]$ at viewpoint $p(n)$. For this experiment, we consider baselines capable of direct novel-view generation at viewpoint $n$, SV4D, GCD, TrajectoryAttention, and TrajectoryCrafter.
As shown in the upper part of Table~\ref{tab:quant_comparison}, Zero4D, despite being training-free, achieves the highest score in subject consistency and ranks second in five other categories. 
This demonstrates that ours achieves robust novel-view video generation performance, comparable to models pretrained on large-scale datasets.

\noindent\textbf{Evaluation of global spatio-temporal consistency.} 
To examine whether models maintain global 4D consistency, we construct re-aligned videos at a fixed viewpoint from generated bullet-time videos.
For each input frame $x[1,i]$ ($i=1,\dots,F$), baselines generate a bullet-time sequence $x[:,i]$ along a predefined trajectory. 
These sequences are aggregated into a 4D grid $x[:,:]$, from which the fixed-view sequence $x[n,:]$ at viewpoint $p(n)$ is extracted. 
We consider three baseline models capable of bullet-time video generation: CameraCtrl, TrajectoryAttention, and TrajectoryCrafter. In contrast, Zero4D directly retrieves $x[n,:]$ from its generated 4D grid without requiring bullet-time re-alignment. 
As shown in the Table~\ref{tab:quant_comparison} (below the horizontal separator), ours achieves the highest scores in five VBench categories and second-best in the remaining two. 
This strong performance indicates that spatio-temporal interpolation enables Zero4D to preserve global consistency across views and time, 
whereas baseline models, unable to sample jointly across multi-view and multi-time dimensions, yield inferior consistency. Although baseline models generate plausible bullet-time results at individual time steps, re-alignment to a fixed viewpoint exposes frequent inconsistencies, particularly in the background and the x–t slices  shown in Figure~\ref{fig:qualitative comparison}, which clearly reveal the inconsistencies.

\subsection{Bullet-time video generation}

We design two evaluations for bullet-time video generation: (1) direct generation along a camera trajectory to assess spatial coherence, and (2) multi-view alignment at fixed time steps to measure global 4D consistency.

\noindent\textbf{Evaluation of direct bullet-time generation.} 
We compare Zero4D against baselines (CameraCtrl, TrajectoryAttention, TrajectoryCrafter) capable of bullet-time generation. Given an input video $x[1,:]$, these models generate bullet-time sequences $x[:,i]$ by smoothly moving the camera along a predefined trajectory at fixed time $i$. This setting provides a direct evaluation of each model’s ability to produce spatially coherent bullet-time videos from the input. As shown in Table~\ref{tab:bullet-table} (upper part), Zero4D attains comparable scores to baselines that are explicitly trained for novel-view video generation, despite being a training-free approach. 

\begin{table}[t]
\centering
\caption{\textbf{Bullet-time video quantitative comparisons.} We report results on (1) direct bullet-time generation for spatial coherence and (2) multi-view consistency by re-aligning outputs at fixed time steps. (* denotes baselines evaluated with novel-view re-alignment)}
\resizebox{\linewidth}{!}{
\begin{tabular}{l|cccc}
    \toprule
    Method & ATE (m, $\downarrow$) & RPE-T ($\downarrow$) & RPE-R (deg $\downarrow$) & MEt3R $\downarrow$ \\
    \midrule
    CameraCtrl~\cite{cameractrl} & 0.185 & 0.155 & 0.57 & \second{0.0264} \\
    TrajectoryAttention~\cite{trajattn} & \second{0.182} & \best{0.113} & \best{0.25} & \best{0.0202} \\
    TrajectoryCrafter~\cite{trajcrafter} & \best{0.170} & \second{0.140} & 2.26 & \second{0.0224} \\
    Ours & 0.190 & 0.142 & \second{0.53} & 0.0307 \\
    \midrule
    TrajectoryAttention*~\cite{trajattn} & 5.582 & 3.377 & \second{1.65} & 0.1000 \\
    TrajectoryCrafter*~\cite{trajcrafter} & \second{0.211} & \second{0.251} & 3.61 & \second{0.0930} \\
    Ours & \best{0.190} & \best{0.142} & \best{0.53} & \best{0.0307} \\
    \bottomrule
\end{tabular}
}
\label{tab:bullet-table}
\end{table}

\noindent\textbf{Evaluation of multi-view consistency in bullet-time.} 
To further assess global 4D consistency, we construct bullet-time videos by re-aligning novel-view outputs at a fixed time step. For baseline models, novel-view videos $x[n,:]$ are generated at each target viewpoint $p(n)$ along the predefined camera trajectory, and the frames corresponding to the same time index are re-aligned to form a bullet-time sequence $x[:,:]$. In contrast, ours directly retrieves the corresponding sequence $x[:,i]$ from its generated 4D grid $x[:,:]$, without requiring re-alignment. As shown in Table~\ref{tab:bullet-table} (below part), Zero4D maintains global coherence across views and time, thereby achieving better accuracy in pose estimation (ATE, RPE-T, RPE-R) and lower MEt3R scores, surpassing the performance of baseline approaches.

\noindent\textbf{Ablation.}
We performed ablation studies under two settings: (1) \textit{Without warped frame guidance:} removing warped frames from the input degrades image fidelity and weakens structural details. (2)\textit{Without spatio-temporal bidirectional interpolation (STBI):} generating each novel-view independently breaks multi-view coherence.  Table~\ref{tab:quant_ablation2}, evaluated with ATE, RPE-T, RPE-R in the bullet-time setting and VBench~\citep{vbench} for fixed novel-view, shows that both components are essential for maintaining fidelity and global consistency. Qualitative ablation results are provided in  Appendix~A.5

\subsection{Limitations}
Our method can generate high-fidelity 4D videos without any training, yet several limitations remain. First, generating wide camera trajectories, such as full 360-degree rotations, remains challenging because monocular video inputs provide limited geometric cues. Second, when the depth estimation model produces inaccurate depth predictions in challenging cases, suboptimal 4D videos may be generated. In addition, since the 4D generation process is guided by the prior knowledge embedded in a pre-trained video diffusion model, our method may also inherit limitations that are commonly observed in generative models.

\vspace{-5pt}
\section{Conclusion}
In this work, we introduced a novel training-free approach for synchronized multi-view video generation using an off-the-shelf video diffusion model. Our method generates high-quality 4D video through depth-based warping and spatio-temporal bidirectional interpolation, ensuring structural consistency across both spatial and temporal domains. Unlike prior methods that rely on extensive training with video or 4D datasets, our framework achieves competitive performance without additional training. Experiments demonstrate that our approach produces synchronized multi-view videos with superior subject consistency, smooth motion trajectories, and temporal stability. This makes our framework a practical solution for multi-view video generation, particularly in scenarios where large-scale 4D datasets and powerful computational resources are limited. Future work may investigate extensions to more complex dynamic scenes, adaptive interpolation strategies, or fusion with other generative models to further enhance realism and flexibility.

{
    \small
    \bibliographystyle{ieeenat_fullname}
    \bibliography{main}
}


\clearpage
\onecolumn
\appendix
\section{Appendix}

\subsection{User study.}
\label{appendix:user_study}
\begin{wraptable}{r}{0.6\linewidth}
    \centering
    \caption{\textbf{User study.} Winning rates across four evaluation metrics. Our method consistently outperforms the baselines, particularly in General Quality and Background Quality.}
    \resizebox{\linewidth}{!}{
    \begin{tabular}{l|cccc}
        \toprule
        \textbf{Method} & \textbf{View Angle} & \textbf{General Quality} & \textbf{Smoothness} & \textbf{BG Quality} \\
        \midrule
        Ours  & \second{30\%}& \best{36\%}& \second{33\%}& \best{39\%}\\
        TrajectoryCrafter& \best{32\%}& \second{30\%}& 27\% & \second{28\%}\\
        TrajectoryAttention& 27\% & 26\% & \best{34\%}& 23\% \\
        CameraCtrl& 11\% & 8\% & 6\% &10\% \\
        \bottomrule
    \end{tabular}
    }
    \label{tab:user_study}
\end{wraptable}

To evaluate our approach, we conducted a user study comparing Ours, TrajectoryCrafter~\citep{trajcrafter}, TrajectoryAttention~\citep{trajattn}, and CameraCtrl~\citep{cameractrl} across four key metrics: View Angle, General Quality, Smoothness, and Background Quality. Participants viewed generated videos and selected the most visually appealing results for each criterion, providing subjective feedback on the overall quality and realism. As shown in Table~\ref{tab:user_study}, our method consistently achieved the highest user preference, particularly excelling in General Quality (36\%) and Background Quality (39\%), which highlights its superior fidelity and ability to preserve scene details. The View Angle metric (30\%) confirms accurate and convincing novel-view synthesis, while Smoothness (33\%) indicates our approach produces fluid transitions with minimal distortion or artifacts. These results collectively demonstrate that our method offers a more immersive and visually coherent experience compared to competing techniques.

\subsection{Pre-trained model checkpoints}
Zero4D is developed based on publicly available, pre-trained generative models for both images and videos. For transparency and reproducibility, we specify below the exact versions of each model employed in our framework:
\begin{itemize}
    \item Depth estimation model: Depthcrafter 
    \item Image-to-Video generation model: stable-video-diffusion-img2vid-xt
\end{itemize}

\subsection{Camera trajectory control}
\label{appendix:camera_control}
\begin{wrapfigure}{r}{0.5\textwidth}
    \vspace{-1.0\baselineskip} 
    \centering
    \includegraphics[width=0.98\linewidth]{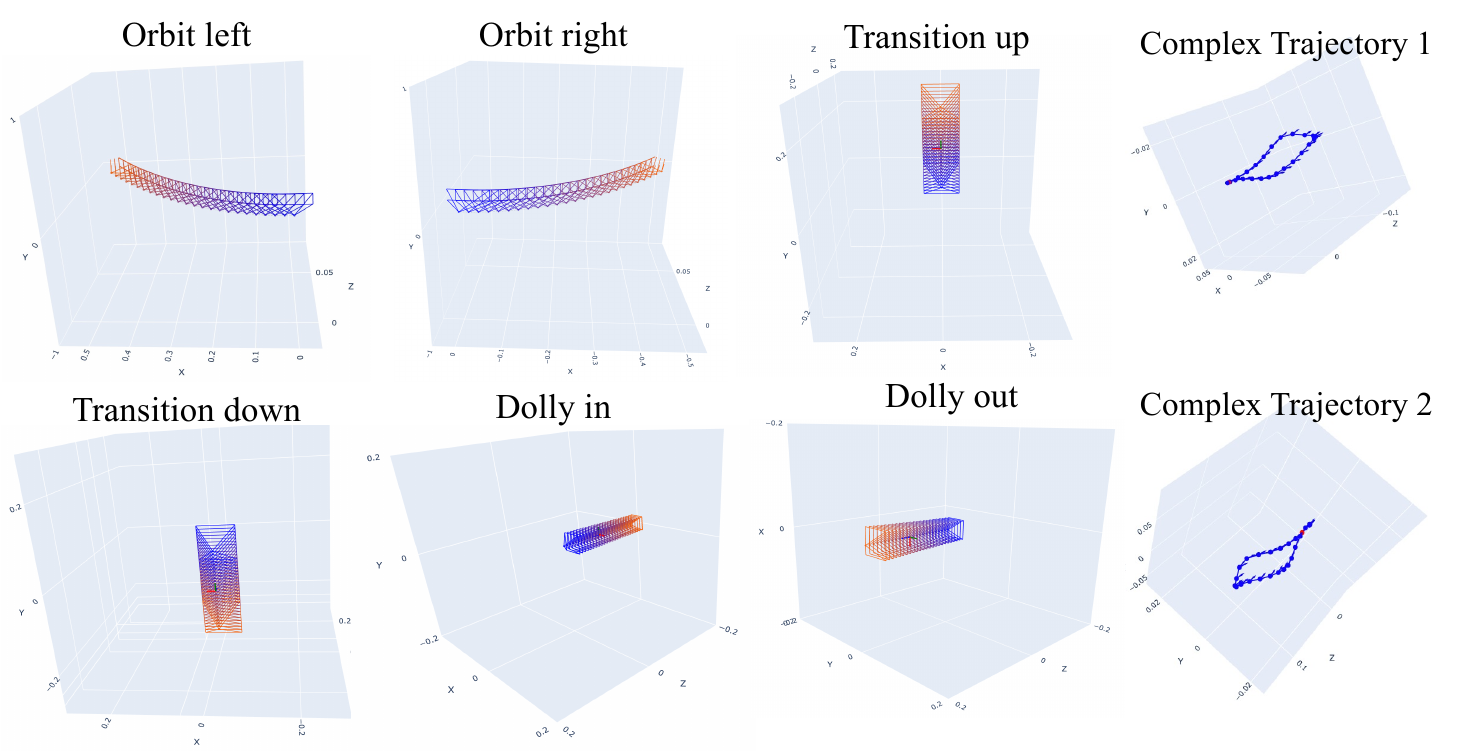}
    \caption{\textbf{Camera trajectory visualization.} With a monocular depth estimation model, our approach can generate various novel view videos with spatio-temporal synchronized videos.}
    \label{fig:warping_camera}
    \vspace{-1.0\baselineskip} 
\end{wrapfigure}

We support various camera motions for novel view synthesis, leveraging depth information for realistic scene transformation:
\newline \textbf{Camera orbit rotation:} Horizontal camera movement around the subject, creating a side-to-side viewing effect. The depth map guides proper parallax by determining each pixel's displacement based on its relative depth.
\newline \textbf{Dolly movement:} Forward/backward camera translation that adjusts focal length to maintain subject size. For dolly-in, foreground elements remain stable while the background compresses; for dolly-out, the background expands naturally.
\newline \textbf{Elevation transition:} Vertical camera movement that rotates the viewpoint up or down. Depth information ensures accurate perspective shifts as the camera changes height, maintaining geometric consistency.
\newline \textbf{Complex trajectory:} We also conducted experiments on complex camera trajectories. In this setting, the camera moves along a combined path in the x, y, and z axes, first moving inward toward the subject and then moving outward, forming a complex trajectory. In addition, we generated another variant, referred to as complex trajectory 2, where the camera first moves outward and then moves back inward.
\newline Our system utilizes monocular video depth estimation to construct a pseudo-3D dynamic representation of the scene. This depth map is crucial for maintaining geometric consistency during novel view synthesis, allowing for convincing parallax effects and occlusion handling. By projecting pixels according to their estimated depth values, we achieve realistic scene transformations without explicit 3D reconstruction. 
\subsection{Details of Zero4D Implementation}
\label{appendix:zero4d_details}

\begin{algorithm}[h]
\LinesNumbered
\caption{$I_\theta$: A sampling step of extended ViBiDSampler for bidirectional interpolation}
\label{alg:bidirectional_sampling}

\SetKwFunction{Itheta}{$I_\theta$}
\SetKwProg{Fn}{Function}{:}{}

\Fn{\Itheta{$x_t, \sigma_t, c_{start}, c_{end}, x_w$}}{

    $\textcolor{orange}{\hat{x}_{c_{start}}} \leftarrow D_{\theta}(x_t;\, \sigma_t, c_{start})$ \tcp*{EDM denoising}

    $\textcolor{red}{\bar{x}_{c_{start}}} \leftarrow 
        \textcolor{orange}{\hat{x}_{c_{start}}}\cdot m 
        + \textcolor{red}{x_w}\cdot (1 - m)$

    $x_{t-1,c_{start}} \leftarrow 
        \textcolor{red}{\bar{x}_{c_{start}}}
        + \frac{\sigma_{t-1}}{\sigma_t}(x_t - \hat{x}_{\emptyset})$

    $(x_t, c_{start}) \leftarrow 
        x_{t-1,c_{start}} +
        \sqrt{\sigma_t^2 - \sigma_{t-1}^2}\,\epsilon$  \tcp*{Re-noise}

    $(x_t, c_{start}) \leftarrow \text{flip}(x_t, c_{start})$ \tcp*{Time reverse}

    $\textcolor{orange}{\hat{x}'_{c_{end}}} \leftarrow 
        D_{\theta}(x'_t, c_{start};\, \sigma_t, c_{end})$ \tcp*{EDM denoising}

    $\textcolor{red}{\bar{x}'_{c_{end}}} \leftarrow 
        \textcolor{orange}{\hat{x}'_{c_{end}}}\cdot m 
        + \textcolor{red}{x_w}\cdot (1 - m)$

    $x'_{t-1} \leftarrow 
        \textcolor{red}{\bar{x}'_{c_{end}}}
        + \frac{\sigma_{t-1}}{\sigma_t}(x'_t - \hat{x}'_{\emptyset})$

    $x'_{t-1} \leftarrow \text{flip}(x'_{t-1})$ \tcp*{Time reverse}
}
\KwRet{$x_{t-1}$}

\end{algorithm}

\begin{algorithm}[h]
\caption{Novel view synthesis and end-view video generation algorithm from \cite{ext_nvs}}
\label{alg:ext_nvs}
\LinesNumbered
\KwIn{Warped frames $x_w$, opacity mask $m$}
\KwOut{Input video $x_0$}

$x_T \sim \mathcal{N}(0,1)$\;

\For{$t \gets T$ \KwTo $1$}{
  \uIf{$t > T - T^{\text{guide}}$}{
    \For{$r \gets 1$ \KwTo $R$}{
      $\hat{x}_0 \leftarrow \text{Predict}(x_t)$\;
      \uIf{$r \leq R^{\text{guide}}$}{
        $\hat{x}_0 \leftarrow D_{\theta}(x_t;\; \sigma_t, c_{x_0})$\;
        $\bar{x}_0 \leftarrow \hat{x}_0 \cdot m + x_w \cdot (1-m)$\;
      }
      \Else{
        $\bar{x}_0 \leftarrow \hat{x}_0$\;
      }
      $x_{t-1} \leftarrow \bar{x}_0 + \frac{\sigma_{t-1}}{\sigma_t}(x_t - \hat{x}_0)$\;
      \If{$r < R$}{
        $x_t \sim \mathcal{N}(\bar{x}_0, \sigma_t)$\;
      }
    }
  }
  \Else{
    $\hat{x}_{t-1} \leftarrow D_{\theta}(x_t;\; \sigma_t, c_{x_0})$\;
    $x_{t-1} \leftarrow \bar{x}_0 + \frac{\sigma_{t-1}}{\sigma_t}(x_t - \hat{x}_0)$\;
  }
}

\Return{$x_0$}\;
\end{algorithm}

\noindent\textbf{Details of interpolation.}
To generate globally consistent 4D videos, we adapt the interpolation strategy during spatio-temporal video generation. Specifically, we leverage ViBiDSampler~\citep{yang2025vibidsampler} as the interpolator $I_{\theta}$. ViBiDSampler is a state-of-the-art training-free video interpolation method designed for image-to-video diffusion models. Given two conditioning frames, it alternates denoising along the temporal axis to synthesize intermediate frames. In our framework, we extend this process by incorporating warped-frame guidance (see Algorithm~\ref{alg:bidirectional_sampling}), which provides additional geometric cues. This modification refines the interpolation process, leading to more faithful structure preservation and improved global spatio-temporal coherence across the generated 4D video grid.

\begin{figure}[t]
    \centering
    \includegraphics[width=0.8\linewidth]{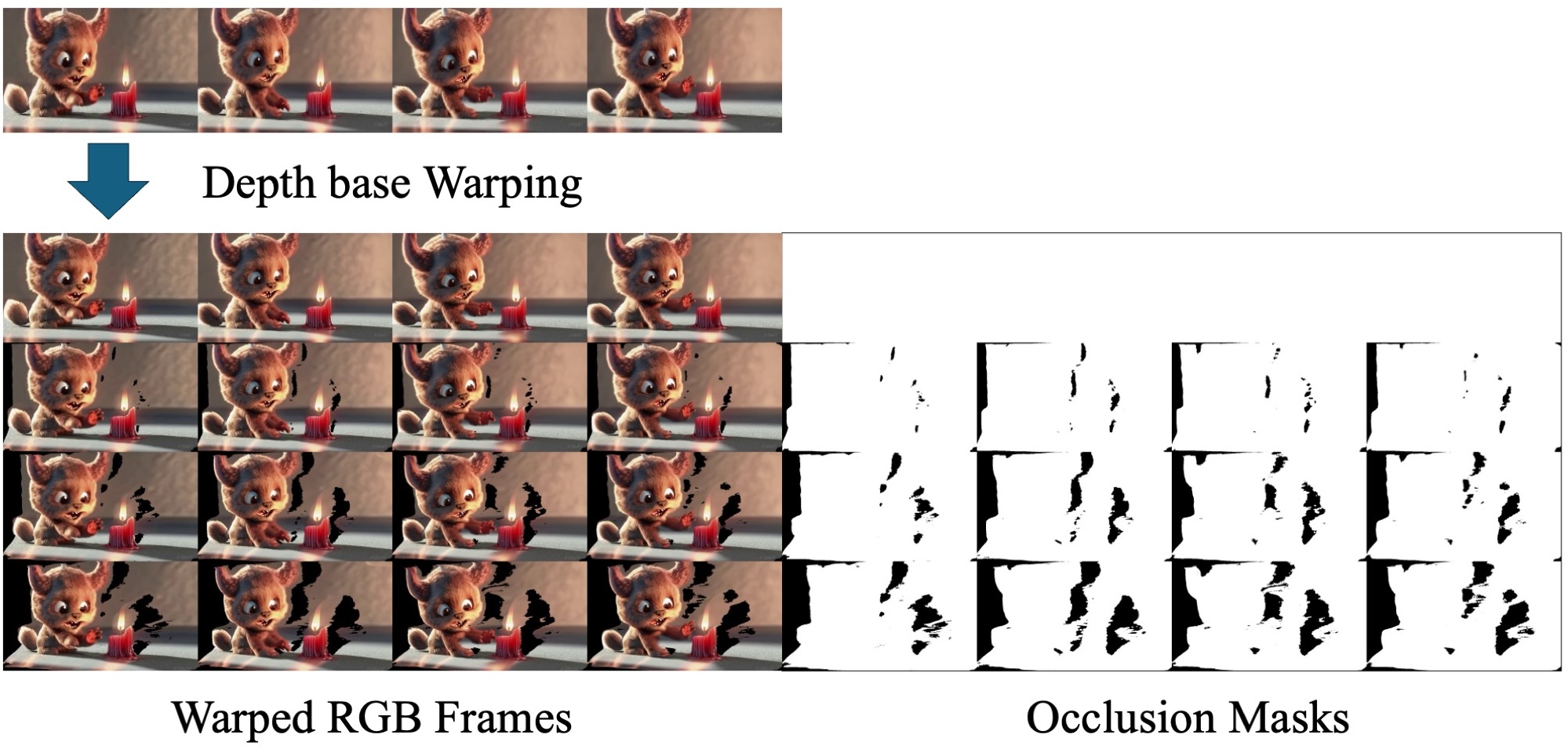}
    \caption{\textbf{Input Video Warping.} Given a single video, we utilize an off-the-shelf depth estimation model to generate warped frames from novel viewpoints.}
    \label{fig:warping_exm}
\end{figure}

\noindent\textbf{Novle-view synthesis.}
Algorithm~\ref{alg:ext_nvs} outlines the process for generating novel-view videos from a single monocular video. We first apply novel view synthesis to the initial frame using an I2V diffusion model \cite{svd} to produce the novel view $x[:,1]$. 
For this, depth-based warping priors from the input video are incorporated to enable inpainting-based synthesis. Specifically, using an off-the-shelf depth estimation model~\cite{depthcrafter}, we warp the original frame to novel viewpoints, as illustrated in Figure~\ref{fig:warping_camera}. As shown in Fig.~\ref{fig:warping_exm}, occluded regions from the warp operation appear black, allowing us to extract an opacity mask. Inspired by~\cite{repaint, nvs_solver, ext_nvs}, we adopt a mask inpainting approach, where inpainting is performed on the estimated noisy frame $\hat{x}_0[:,1]$. Rather than applying inpainting at every denoising step, as in~\cite{ext_nvs}, we utilize a re-noising process within the diffusion model’s denoising step to refine the final synthesis by reducing artifacts and enhancing structural coherence. A detailed description is provided in Algorithm~\ref{alg:ext_nvs}.

\subsection{Additional Results}
\label{appendix:additional_results}

\begin{figure*}[ht]
  \centering
  \begin{minipage}[b]{0.003\linewidth}
  \centering
   \rotatebox{90}{\fontsize{8}{9}\selectfont w/o STBI}\\[1em]
   \vspace{-0.25em} 
   \rotatebox{90}{\fontsize{8}{9}\selectfont w/o warp}\\[1em]
   \vspace{0.2em} 
   \rotatebox{90}{\fontsize{8}{9}\selectfont Ours}
   \vspace{0.2em} 
  \end{minipage}
  \hspace{0.005\linewidth} 
  \begin{minipage}[b]{0.98\linewidth}
    \includegraphics[width=\linewidth]{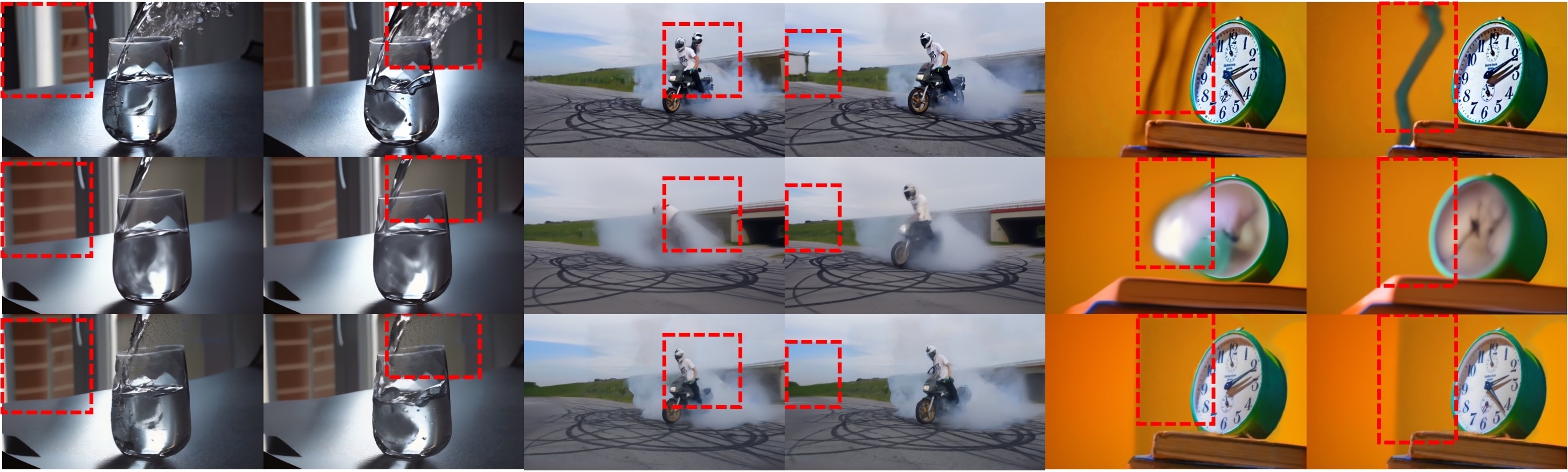}
  \end{minipage}
  \caption{\textbf{Ablation results.} Removing spatio-temporal bidirectional interpolation (STBI) or warping guidance leads to broken consistency and geometric artifacts (red boxes). In contrast, our full method preserves spatial structure and temporal coherence across views.}
  \label{fig:main_ablation}
\end{figure*}

\noindent\textbf{Ablation (detailed analysis).}
Figure~\ref{fig:main_ablation} qualitatively illustrates the role of each component in maintaining global consistency. Without spatio-temporal bidirectional interpolation (STBI), each frame is synthesized independently, which causes temporal flickering and background inconsistencies across views. For example, in the water-pouring sequence (left), the liquid surface fails to remain temporally stable, as highlighted by the red boxes. Similarly, without warping guidance, the model struggles with geometric alignment. In the motorcycle example (middle), artifacts appear in the generated human figure, leading to distorted or incomplete shapes. Finally, in the clock sequence (right), the absence of warping or spatio-temporal interpolation leads to visible structural mismatches and background inconsistencies. In contrast, our full model effectively aggregates global information through STBI and enforces geometric consistency via warped-frame guidance, resulting in coherent and high-quality multi-view videos across both spatial and temporal dimensions.

\noindent\textbf{Quantitative comparison under difficult scenarios.} 
We evaluated our method under difficult scenarios (low-light, blurred camera) to evaluate the robustness of our method.
We evaluated the robustness of our proposed method under challenging scenarios, including low-light conditions and camera motion-blurred settings. We use 50 WebVid-10M videos with a low-light filter(30$\%$ brightness) and a motion blur filter (20px, 45°). As shown in Table~\ref{tab:input_video_perturb}, despite minor drops in aesthetic quality, performance remains stable, especially in bullet-time, where static objects are well preserved.

\begin{table*}[t]
    \centering
    \caption{\textbf{Quantitative results under low-light and camera-motion blur.} We evaluate our method in challenging scenarios where the depth estimation model may fail. We perturb the input videos using motion-blur and low-light filters. The top three rows show bullet-time videos(*), while the bottom three rows present novel-view videos.}
    \resizebox{1.0\linewidth}{!}{
    \begin{tabular}{l|c c c c c c c}
        \toprule
        \textbf{Method} & 
        \makecell[c]{Subject\\Consistency} $\uparrow$ & 
        \makecell[c]{Background\\Consistency} $\uparrow$ & 
        \makecell[c]{Temporal\\Flickering} $\uparrow$ & 
        \makecell[c]{Motion\\Smoothness} $\uparrow$ &
        \makecell[c]{Dynamic\\Degree} $\downarrow$& 
        \makecell[c]{Image\\Quality} $\uparrow$ & 
        \makecell[c]{Aesthetic\\Quality} $\uparrow$ \\
        \midrule
        Ours (Low-light)*& \second{95.03\%}& \best{95.41\%}& \best{98.80\%}& \best{99.31\%}& \second{2.00\%}& 34.12\%& \second{33.46\%}\\
        Ours (Motion blur)*& 94.62\%& 92.88\%& 94.28\%& 94.32\%& \best{2.22\%}& \second{37.75\%}& 28.81\\
        Ours *& \best{95.73\%}& \second{94.81\%}& \second{96.88\%}& \second{98.76\%}& 1.00\%& \best{38.81\%}& \best{38.14\%}\\
        \bottomrule
 Ours (Low-light)& 94.28\%& \second{96.03\%}& \second{95.98\%}& \best{99.22\%}& \second{30.13\%}& 36.11\%&\second{33.76\%}\\
 Ours (Motion blur)& \second{94.76\%}& 94.68\%& 95.49\%& 95.99\%& 32.35\%& \second{39.11\%}&29.21\\
 Ours& \best{95.55\%} & \best{95.75\%} & \best{97.48\%}& \second{98.34\%}& \best{27.50\%}& \best{51.12\%} &\best{38.22\%}\\
        \bottomrule
    \end{tabular}
    }
    \label{tab:input_video_perturb}
\end{table*}

\end{document}